\documentclass{article}

\usepackage{amsmath,amsfonts,bm}

\def\eqref#1{equation~\ref{#1}}

\def\1{\bm{1}}

\DeclareMathAlphabet{\mathsfit}{\encodingdefault}{\sfdefault}{m}{sl}
\SetMathAlphabet{\mathsfit}{bold}{\encodingdefault}{\sfdefault}{bx}{n}

\newcommand{\E}{\mathbb{E}}

\usepackage{microtype}
\usepackage{graphicx}
\usepackage{subfigure}
\usepackage{booktabs} 
\usepackage{mathtools}
\usepackage{amsfonts}
\usepackage{xfrac}
\usepackage[nice]{nicefrac}
\usepackage{url}
\usepackage{hyperref}

\newcommand{\alglinelabel}{%
  \addtocounter{ALC@line}{-1}
  \refstepcounter{ALC@line}
  \label
}

\newcommand{\meqref}{\ref}
\newcommand{\onetau}{\one_\tau}

\usepackage{pgfplots}
\usepgfplotslibrary{groupplots}
\newcommand{\scplotwidth}{0.45\columnwidth}
\newcommand{\scplotheight}{0.31\columnwidth}

\newcommand{\plotwidth}{0.73\columnwidth}
\newcommand{\plotheight}{0.50\columnwidth}

\pgfplotsset{width=\plotwidth, height=\plotheight,compat=1.9}

\newcommand\mtlarge{\fontsize{8pt}{10pt}\selectfont}
\newcommand\markSize{0.8}
\newcommand\pointInt{1}
\pgfplotsset{every axis/.append style={
xlabel={$x$},          
ylabel={$y$},          
label style={font=\mtlarge},
tick label style={font=\mtlarge},
legend style={font=\mtlarge}
}}

\newcommand{\bse}{\begin{subequations}}
\newcommand{\ese}{\end{subequations}}

\ifodd 0
    
    \newcommand{\mycom}[1]{\textbf{\color{red} (COMMENT: #1)}} 
\else
    
    \newcommand{\mycom}[1]{}
\fi

\newcommand\Hil{\mathcal{H}}
\newcommand\leb{\left(}
\newcommand\rib{\right)}

\newcommand\better{\succ}

\newcommand\one{\mathbf{1}}
\newcommand{\Prob}{\mathbb{P}}
\newcommand{\covN}{\mathcal{N}(\mathcal{B}_f,\epsilon,\|\cdot\|_\infty)}
\newcommand\tf{f}

\newcommand\hh{\tilde{h}}
\newcommand\kk{\tilde{k}}
\newcommand\dd{\tilde{d}}

\newcommand\xx{\tilde{x}}
\newcommand\ff{\tilde{f}}

\newcommand{\be}{\begin{equation}}
\newcommand{\ee}{\end{equation}}
\newcommand\norm[1]{\left\lVert#1\right\rVert}

\newcommand{\defeq}{\vcentcolon=}

\usepackage{xcolor}
\usepackage{pifont}

\allowdisplaybreaks[4]

\ifodd 0
    \newcommand\rev[1]{{\color{blue}#1}}
    \newcommand{\com}[1]{\textbf{\color{red} (COMMENT: #1)}} 
\else 
    \newcommand{\rev}[1]{{#1}}
    \newcommand{\com}[1]{}
\fi

\def\abs#1{|{#1}|}

\def\meqref#1{(\ref{#1})}

\def\citep#1{(\cite{#1})}

\usepackage{hyperref}

\usepackage{mathtools}
\usepackage{environ}

\NewEnviron{resizealigned}{\sbox0{
    $\begin{matrix}\displaystyle\BODY\end{matrix}$}%
  \sbox1{$(\theequation)$}%
  \sbox2{\parbox{\dimexpr \wd0 + 2\wd1}%
    {\begin{equation}\begin{aligned}\BODY\end{aligned}\end{equation}}}
  \resizebox{\columnwidth}{!}{\usebox2}%
}

\usepackage[accepted]{icml2024}

\usepackage{amsmath}
\usepackage{amssymb}
\usepackage{amsthm}

\usepackage[capitalize,noabbrev]{cleveref}

\theoremstyle{plain}
\newtheorem{theorem}{Theorem}[section]

\newtheorem{lemma}[theorem]{Lemma}

\theoremstyle{definition}

\newtheorem{assump}[theorem]{Assumption}
\theoremstyle{remark}
\newtheorem{remark}[theorem]{Remark}

\usepackage[textsize=tiny]{todonotes}

\icmltitlerunning{Principled Preferential Bayesian Optimization}

\begin{document}

\twocolumn[
\icmltitle{Principled Preferential Bayesian Optimization}



\icmlsetsymbol{equal}{*}

\begin{icmlauthorlist}
\icmlauthor{Wenjie Xu}{epfl,empa}
\icmlauthor{Wenbin Wang}{epfl}
\icmlauthor{Yuning Jiang}{epfl}
\icmlauthor{Bratislav Svetozarevic}{empa,ai_inst}
\icmlauthor{Colin N. Jones}{epfl}
\end{icmlauthorlist}

\icmlaffiliation{epfl}{Automatic Control Laboratory, EPFL, Lausanne, Switzerland}
\icmlaffiliation{empa}{Urban Energy Systems Laboratory, Empa, Zurich, Switzerland}
\icmlaffiliation{ai_inst}{The Institute for Artificial Intelligence Research and Development of Serbia, Serbia}

\icmlcorrespondingauthor{Yuning Jiang}{yuning.jiang@ieee.org}

\icmlkeywords{Machine Learning, ICML}

\vskip 0.3in
]



\printAffiliationsAndNotice{}  

\begin{abstract}
We study the problem of preferential Bayesian optimization~(BO), where we aim to optimize a black-box function with only \emph{preference} feedback over a pair of candidate solutions. Inspired by the \emph{likelihood ratio} idea, we construct a confidence set of the black-box function using only the preference feedback. An optimistic algorithm with an efficient computational method is then developed to solve the problem, which enjoys an information-theoretic bound on the \rev{total} cumulative regret, a \emph{first-of-its-kind} for preferential BO. This bound further allows us to design a scheme to report an estimated best solution, with a guaranteed convergence rate. Experimental results on sampled instances from Gaussian processes, standard test functions, and a thermal comfort optimization problem all show that our method stably achieves better or competitive performance as compared to the existing state-of-the-art heuristics, which, however, do not have theoretical guarantees on regret bounds or convergence.      
\end{abstract}

\section{Introduction}
\label{sec:intro}
Bayesian optimization (BO) is a popular sample-efficient black-box optimization method~\cite{shahriari2015taking,frazier2018tutorial}. It is widely applied to tuning hyperparameters of machine learning models~\citep{snoek2012practical}, optimizing the performance of control systems~\citep{xu2022vabo}, and discovering new drugs~\citep{negoescu2011knowledge}, etc. 

The main idea of BO is based on \emph{surrogate modeling}. That is, a learning algorithm~(typically Gaussian process regression) is applied to learn the unknown black-box function using historical samples, which then outputs a learned surrogate together with uncertainty quantification. Then BO algorithms, such as the popular Expected Improvement~\cite{jones1998efficient} and GP-UCB algorithms~\cite{srinivas2012information}, use the information of this learned surrogate and uncertainty quantification to choose the next sample point. 

The conventional BO setting assumes each sample, which typically corresponds to a round of real-world experiment or software simulation in practice, returns a noisy scalar evaluation of the black-box function. However, many \emph{human-in-the-loop} systems can not return such a scalar value, or it is much more difficult to directly obtain such a scalar evaluation from humans since humans are bad at sensing absolute magnitude~\cite{kahneman2013prospect}. In contrast, it is much easier for a human to compare a pair of solutions and report which is preferred~\rev{\cite{lichtenstein1971reversals,tversky1974judgment,kahneman2013prospect}}.    

This gives rise to \emph{preferential Bayesian optimization}~\cite{gonzalez2017preferential}, where the scalar evaluation of the black-box function is not available. But rather, we can query an oracle to compare a pair of solutions, or the so-called \emph{duels}. Such settings arise widely in a broad range of applications, such as visual design optimization~\cite{koyama2020sequential}, thermal comfort optimization~\cite{abdelrahman2022targeting} and robotic gait optimization~\cite{li2021roial}. 

Existing preferential Bayesian optimization methods are mostly heuristic, without formal guarantees on cumulative regret or convergence to the global optimal solution. For example, \cite{gonzalez2017preferential} proposes several heuristic acquisition strategies, including expected improvement and Thompson sampling-based methods, for preferential Bayesian optimization. \cite{mikkola2020projective} extends the preferential Bayesian optimization to the projective setting. \cite{pmlr-v202-takeno23b} proposes a Thompson sampling-based method for practical preferential Bayesian optimization with skew Gaussian process. \cite{astudillo2023qeubo} proposes a decision theoretical acquisition strategy with a convergence rate guarantee for a finite input set. However, as far as we know, all the existing preferential Bayesian optimization methods can not provide theoretical guarantees on cumulative regret or global convergence with continuous input space, partially due to the challenge of quantifying uncertainty in a principled way.  

Beyond preferential BO, optimization from preference feedback has also been investigated in other contexts. In the following, we first survey the related work other than preferential BO and then highlight our unique contributions.    

\textbf{Dueling Bandits} In dueling bandits~\cite{yue2012k}, the goal is to identify the best arm from a set of finite arms, using only the noisy comparison feedback. It has also been extended to adversarial~\cite{gajane2015relative} and contextual~\cite{dudik2015contextual,saha2022efficient} settings. One extension that is most related to this work is kernelized dueling bandits~\cite{sui2017multi,sui2018stagewise}. However, this line of research is typically restricted to the case where the number of arms is finite, and the regret bound can blow up to infinity when the number of arms goes to infinity~(e.g., Thm.~2 in~\cite{sui2017multi}). \rev{A recent work~\cite{mehta2023kernelized} proposes an offline method with suboptimality bound by learning winning probability, which, however, are not applicable to online learning problems due to linear growth of regret over the randomly sampled compared point sequences.} In the existing literature, there is no \rev{\emph{cumulative}} regret bound that depends on an inherent complexity metric~(such as covering number and maximum information gain~\cite{srinivas2012information}) of the black-box function with continuous input space. 

\textbf{Convex Optimization with Preference Feedback} \cite{saha2021dueling, yue2009interactively} consider the optimization of convex functions, where only a comparison oracle of function values over different points is available. The proposed methods estimate the gradient from the preference signals. However, this line of research restricts the function to be convex, while in practice, the black-box function may be non-convex. The proposed method may get stuck in a local optimum and can be sample-inefficient since each estimate of the gradient already needs several samples.      

\textbf{Reinforcement Learning from Human Feedback} Reinforcement learning from human feedback (RLHF)~\cite{christiano2017deep,griffith2013policy} has recently become very popular. It has found many successes in wide applications, including training robots~\cite{hiranaka2023primitive}, playing games~\cite{warnell2018deep}, and remarkably large language models~\cite{ouyang2022training}. On the theoretical line of RLHF research, recent results analyze the offline learning of the implicit reward function~\cite{pmlr-v202-zhu23f} and the model-based optimistic reinforcement learning from human feedback~\cite{wang2023rlhf}. However, the existing theoretical analysis either only deals with finite-dimensional generalized linear models or highly relies on the complexity measure of Eluder dimension~\cite{osband2014model}. The existing generic theoretical analysis for RLHF can not be directly applied to the Bayesian optimization setting, where the Eluder dimension of the infinite-dimensional reproducing kernel Hilbert space is not well understood.   

\textbf{Optimistic Model-based Sequential Decision Making}
Optimism in the face of uncertainty is a widely adopted design principle for model-based sequential decision making problems, such as in Bayesian optimization/reinforcement learning~\cite{wu2022bayesian,xu2023constrained,pacchiano2021towards,curi2020efficient,liu2023optimistic}. The optimism principle has also been applied to RLHF~\cite{wang2023rlhf} recently. However, as far as we know, there is no existing principled optimistic algorithm for preferential BO yet. 

\textbf{Our contributions}. Guided by the optimism principle, we design a preferential Bayesian optimization algorithm that enjoys information-theoretic bounds on the cumulative regret. Specifically, our contributions include:
\vspace{-2ex}
\begin{itemize}
     \setlength\itemsep{0.05em}
    \item \textbf{Algorithm design.} Inspired by the recent work of the confidence set based on {optimistic maximum likelihood estimate}~\cite{liu2023optimistic} and the \emph{likelihood ratio} confidence set idea~\cite{owen1990empirical,emmenegger2023likelihood}, we construct a confidence set by only using the preference feedback. We then exploit the principle of optimism in the face of uncertainty to design a \textbf{P}rincipled \textbf{O}ptimistic \textbf{P}referential \textbf{B}ayesian \textbf{O}ptimization (\textbf{POP-BO}) algorithm, together with a scheme of reporting an estimated best solution.      
    \item \textbf{Theoretical analysis.} Under some mild regularity assumptions, we prove an information-theoretic bound on the cumulative regret of {POP-BO} algorithm, which is \emph{first-of-its-kind}~\footnote{\rev{\cite{mehta2023kernelized} provides a bound on the partial cumulative regret, which only captures the suboptimality of one point in each compared duel. We consider stronger total cumulative regret over both points in the compared duel. See Appendix~\ref{sec:add_contri} for a detailed discussion.}} for preferential Bayesian optimization. This is significant since previous information-theoretic regret bounds typically assume the direct scalar evaluations of black-box functions~\cite{srinivas2012information} while the recent generic theoretical results for RLHF typically rely on Eluder dimension, which is not well understood for RKHS.       
    \item \textbf{Efficient computations.} The optimistic algorithm needs to solve bi-level optimization problems with the inner variable in an infinite-dimensional function space. We leverage the \emph{representer theorem}~\cite{scholkopf2001generalized} to reduce the inner optimization problem to finite-dimensional space, which turns out to be tractable via convex optimization. This further allows \rev{efficient grid-free joint optimization}. 

    \item \textbf{Empirical validations and toolbox.~\footnote{Code link: \href{https://github.com/PREDICT-EPFL/POP-BO}{https://github.com/PREDICT-EPFL/POP-BO}}} Experimental results show that POP-BO consistently achieves better or competitive performance as compared to the state-of-the-art heuristic baselines and more than $10$ times speed-up in computation as compared to the Thompson sampling based method. We also provide a reusable toolbox for future applications of our method.
\end{itemize} 

\section{Problem Statement}

We consider the maximization of a black-box function $f$,
\begin{equation}
    \max_{x\in\mathcal{X}}\;f(x),
\end{equation}
where $\mathcal{X}\subset\mathbb{R}^d$ with $d$ as the input dimension. We use $x\better x^\prime$ to denote the event that `$x$ is preferred to $x^\prime$'. In contrast to the standard BO setup, we assume that we can not directly evaluate the scalar value of $f(x)$ but rather, we have a comparison oracle that compares any two points $x, x^\prime$ and returns a preference signal $\mathbf{1}_{x\better x^\prime}$, which is defined as 

\begin{equation}
\one_{x\better x^\prime}=
    \begin{cases}
        1, & \text{ if } x \text{ is preferred,}\\
        0, & \text{ if } x^\prime \text{ is preferred.}
    \end{cases}
\end{equation}


Before proceeding, we state a set of common assumptions.  

\begin{assump}\label{assump:support_set}
$\mathcal{X}$ is compact and nonempty.
\end{assump}
Assumption~\ref{assump:support_set} is reasonable because, in many applications~(e.g., continuous hyperparameter tuning) of Bayesian Optimization, we are able to restrict the optimization into certain ranges {based on domain knowledge}.
Regarding the black-box function $f$, we assume that,
\begin{assump}
\label{assump:bounded_norm}
$f\in\Hil_{k}$, where $k:\mathbb{R}^d\times\mathbb{R}^d\to \mathbb{R}$ is a symmetric, positive semidefinite kernel function and $\Hil_{k}$ is the corresponding reproducing kernel Hilbert space~(RKHS, see~\cite{scholkopf2001generalized}). Furthermore, we assume $\|f\|_{k}\leq B$, where $\|\cdot\|_{k}$ is the norm induced by the inner product in the corresponding RKHS. 
\end{assump}

Assumption~\ref{assump:bounded_norm} requires that the function to be optimized is {regular} in the sense that it has a bounded norm in the RKHS, which is a common assumption~\cite{chowdhury2017kernelized,zhou2022kernelized}. For simplicity, we will use $\mathcal{B}_f$ to denote the set $\left\{\ff\in\Hil_{k}|\|{\ff}\|_{k}\leq B\right\}$, which is a ball with radius $B$ in $\mathcal{H}_k$. 
\begin{remark}[\textbf{Choice of $B$}]\label{remark_choice_B}
  In practice, a tight norm bound $B$ might not be known beforehand. In the theoretical analysis, we only assume that there is a finite bound $B$, possibly unknown beforehand. In the practical implementation of our algorithm, we can adapt $B$ based on hypothesis testing~\cite{newey1994large}. For example, we can double $B$ every time we detect a low likelihood value~\rev{(See more elaboration in Appendix~\ref{app_sec:ela_on_hyp}.)}. 
\end{remark}

\begin{assump}
\label{assump:kernel_bound}
$k(x,x^\prime)\leq 1, \forall x,x^\prime\in \mathcal{X}$ and $k(x, x^\prime)$ is continuous on $\mathbb{R}^d\times \mathbb{R}^d$. 
\end{assump}
Assumption~\ref{assump:kernel_bound} is a commonly adopted mild assumption in the BO literature~\cite{srinivas2012information,chowdhury2017kernelized}. It holds for most commonly used kernel functions after normalization, such as the linear kernel, the Mat\'ern kernel, and the squared exponential kernel.   

\begin{assump}
\label{assump:pref_model}
    The random preference feedback $\one_{x\better x^\prime}$ from the comparison oracle follows the Bernoulli distribution with $\Prob(\one_{x\better x^\prime}=1)=p_{x\better x^\prime}=\sigma(y-y^\prime)$, where $y=f(x)$, $y^\prime=f(x^\prime)$ and $\sigma(u)=\nicefrac{1}{(1+e^{-u})}$. 
 \end{assump}
Assumption~\ref{assump:pref_model} equivalently assumes that, 
\begin{equation}
\Prob(\one_{x\better x^\prime}=1)=\frac{e^{f(x)}}{e^{f(x)}+e^{f(x^\prime)}},
\end{equation}
which can be observed to be the widely used Bradley-Terry-Luce~(BTL) model~\cite{bradley1952rank} for pairwise comparison. The intuition here is that the more advantage $f(x)$ has as compared to $f(x^\prime)$, the more likely $x$ is preferred. The same comparison model is also used in, e.g., training large language models~\cite{ouyang2022training}.
 At step $t$, our algorithm queries the pair $(x_t, x_t^\prime)$ and the comparison oracle returns the random preference $\mathbf{1}_{x_t\better x_t^\prime}\in\{0,1\}$. For the simplicity of notation, we use $\onetau\in\{0,1\}$ to denote the realization of the Bernoulli random variable $\mathbf{1}_{x_\tau\better x_\tau^\prime}$ when querying the comparison oracle at step $\tau$.  Based on the historical comparison results 
\begin{equation}
    \mathcal{D}_t\defeq\{(x_\tau, x_\tau^\prime, \onetau)\}_{\tau=1}^t,
\end{equation}
the algorithm needs to decide the next pair of samples to compare. Without further notice, all the theoretical results in this paper are under the assumptions~\ref{assump:support_set},~\ref{assump:bounded_norm},~\ref{assump:kernel_bound},~\ref{assump:pref_model}, and all the corresponding proofs are in the appendices.     

\section{High Confidence Set}
\label{sec:high_conf_set}
\textbf{Notations}. The probability, denoted as $\Prob(\cdot)$, is taken over the randomness of the preference feedback generated by the comparison oracle and the randomness generated by the algorithm. Let the filtration $\mathcal{F}_t$ capture all the randomness up to step $t$. $\mathcal{N}(\mathcal{B}_f, \epsilon, \|\cdot\|_\infty)$ denotes the standard covering number~\cite{zhou2002covering} of the function space ball $\mathcal{B}_f$ with the covering balls' radius $\epsilon$ and the infinity norm $\|\cdot\|_\infty$. We will also use $[\tau]$ to denote the set $\{1,\cdots, \tau\}$.  

\subsection{Likelihood-based Confidence Set}
We first introduce the function,
\begin{align}
    p_{\hat{f}}(x_\tau, x_\tau^\prime, \onetau)\defeq& \onetau\sigma(\hat{f}(x_\tau)-\hat{f}(x_\tau^\prime))+\\
    &(1-\one_\tau)\left(1-\sigma(\hat{f}(x_\tau)-\hat{f}(x_\tau^\prime))\right)\nonumber,
\end{align}
    which is the likelihood of $\hat{f}$ over the event $\one_{x_\tau\better x_\tau^\prime}=\one_\tau$ under the Bernoulli preference model in Assumption~\ref{assump:pref_model}.
    
We can then derive the likelihood function of a fixed function $\hat{f}$ over the historical preference dataset $\mathcal{D}_t$~\footnote{Note that $\Prob_{\hat{f}}(\cdot)$ is the likelihood function in $\hat{f}$ over the historical data $\mathcal{D}_t$, not the probability taken over the data/algorithm randomness.}. 
\begin{equation}
\Prob_{\hat{f}}((x_\tau, x_\tau^\prime,\onetau)_{\tau=1}^t)\defeq \prod_{\tau=1}^tp_{\hat{f}}(x_\tau, x_\tau^\prime,\onetau) 
\end{equation}
Taking log gives the log-likelihood function, 
\begin{align}
\ell_t(\hat{f})\defeq&\log\Prob_{\hat{f}}((x_\tau, x_\tau^\prime,\onetau)_{\tau=1}^t)= \sum_{\tau=1}^t\log p_{\hat{f}}(x_\tau, x_\tau^\prime,\onetau)\nonumber  \\
=&\sum_{\tau=1}^t\log\left(\frac{e^{z_\tau}\one_{\tau}+e^{z_\tau^\prime}(1-\one_\tau)}{e^{z_\tau}+e^{z_\tau^\prime}}\right)\\
=&\sum_{\tau=1}^t \left(z_\tau\one_\tau+z_\tau^\prime(1-\one_\tau)\right)-\sum_{\tau=1}^t\log\left(e^{z_\tau}+e^{z_\tau^\prime}\right),\nonumber
\end{align}
where $z_\tau=\hat{f}(x_\tau),z_\tau^\prime=\hat{f}(x_\tau^\prime)$, $\onetau\in\{0,1\}$ is the data realization of $\one_{x_\tau\better x_\tau^\prime}$, and the last equality can be checked correct for either $\onetau=1$ or $\onetau=0$.

A common method for statistical estimation is by maximizing the likelihood. Hence, we introduce the maximum likelihood estimator~(MLE), 
\begin{equation}
\hat{f}^\mathrm{MLE}_{t}\in\arg\max_{\ff\in\mathcal{B}_f}\log\Prob_{\ff}((x_\tau, x_\tau^\prime,\onetau)_{\tau=1}^{t}).
\end{equation}

With the maximum likelihood estimator introduced, the posterior high confidence set can be derived as shown in Thm.~\ref{thm:conf_set_f} using the maximum log-likelihood value.  

\begin{theorem}[\textbf{Likelihood-based Confidence Set}]
\label{thm:conf_set_f}
$\forall \epsilon, \delta>0$, let,
\begin{equation}
    \mathcal{B}^{t+1}_f\defeq\{\tilde{f}\in\mathcal{B}_f|\ell_t(\ff)\geq\ell_t(\hat{f}^{\mathrm{MLE}}_t)-\beta_1(\epsilon, \delta, t)\},
\end{equation}
where $\beta_1(\epsilon, \delta, t)\defeq\sqrt{32tB^2\log\frac{\pi^2t^2\covN}{6\delta}}+C_L\epsilon t=\mathcal{O}\left(\sqrt{t\log\frac{t\mathcal{N}(\mathcal{B}_f, \epsilon,\|\cdot\|_\infty)}{\delta}}+\epsilon t\right)$, with $C_L$ a constant independent of $\delta, t$ and $\epsilon$.
We have, 
\begin{equation}
\Prob\left(\tf\in\mathcal{B}^{t+1}_f,\forall t\geq1\right)\geq1-\delta.
\end{equation}
\end{theorem}
\begin{figure*}[htbp!]
    \centering
   {
    \includegraphics[width=2.1\columnwidth]{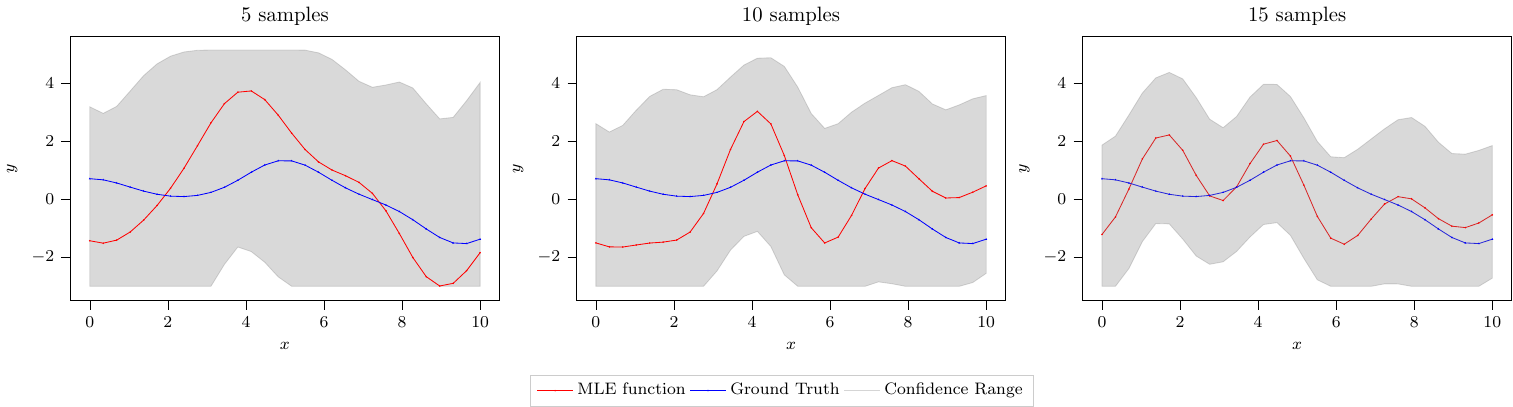}
    }
    \caption{Demonstration of the maximum likelihood function and the confidence set based on likelihood. The results are derived using random sequential comparisons~(that is, comparing $x_t$ to $x_{t-1}$), where each $x_t$ is uniformly randomly sampled from the input set.}
    \label{fig:conf_set_demo}
\end{figure*}
Intuitively, the confidence set $\mathcal{B}_f^{t+1}$ includes the functions with the log-likelihood value that is only `a little worse' than the maximum likelihood estimator. It turns out that by correctly setting the `worse' level $\beta_1$, the confidence set $\mathcal{B}_f^{t+1}$ contains the ground-truth function $f$ with high probability. This is reasonable because the preference data is generated with the ground-truth function, and thus the likelihood of the ground-truth function will not be too much lower than the maximum likelihood estimator. 

\begin{remark}[\textbf{Choice of $\epsilon$}]
In Thm.~\ref{thm:conf_set_f}, $\beta_1(\epsilon, \delta, t)$ also depends on a small positive value $\epsilon$, which is to be chosen. In the theoretical analysis, it will be seen that $\epsilon$ can be selected to be $\nicefrac{1}{T}$, where $T$ is the algorithm's running horizon.  \end{remark}

\begin{remark}[\textbf{Likelihood Ratio Idea}]
The confidence set $\mathcal{B}_f^{t+1}$ contains the functions $\ff$ that satisfy,
\begin{equation}
\frac{\Prob_{\tilde{f}}((x_\tau, x_\tau^\prime,\onetau)_{\tau=1}^t)}{\Prob_{\hat{f}^\mathrm{MLE}_{t}}((x_\tau, x_\tau^\prime,\onetau)_{\tau=1}^t)}\geq e^{-\beta_1(\epsilon, \delta, t)},
\end{equation}
which is the likelihood ratio confidence set~\cite{owen1990empirical}.
\end{remark}
\begin{remark} 
\rev{Surrogate-based black-box optimization with kernel method is often referred to as Bayesian optimization
due to its close relations to Bayesian Gaussian process model.
Hence, we refer to our method as preferential BO.}
\end{remark}
Based on the confidence set in Thm.~\ref{thm:conf_set_f}, we can derive the pointwise confidence range for the black-box function.
\begin{equation}
  \inf_{\ff\in\mathcal{B}^t_f}\ff(x)\leq f(x)\leq \sup_{\ff\in\mathcal{B}^t_f}\ff(x).  
\end{equation}
Fig.~\ref{fig:conf_set_demo} demonstrates the maximum likelihood estimate function and the confidence range with the ground truth function sampled from a Gaussian process, random comparison inputs, and $\beta_1(\epsilon, \delta, t)$ set to be a constant $1.0$. It can be seen that the maximum likelihood estimate approximates the ground truth better and better with the confidence range shrinking, as we have more and more comparison data.





\subsection{Bound Duel-wise Error}
\label{sec:aug_RKHS}
Thm.~\ref{thm:conf_set_f} gives a high confidence set based on the likelihood function. However, it is not straightforward how the likelihood bounds lead to the error bounds on function value differences over a compared pair $(x, x^\prime)$, which determines the preference distribution. The following theorem further gives such a bound over the historical samples.      
\begin{lemma}[\textbf{Elliptical Bound}]\label{thm:obj_ell_po} For any estimate $\hat{f}_{t+1}\in\mathcal{B}_f^{t+1}$ that is measurable with respect to the filtration $\mathcal{F}_{t}$, we have, with probability at least $1-{\delta}$, $\forall t\geq1$, 
\begin{align}
  \sum_{\tau=1}^t&\left(\left(\hat{f}_{t+1}(x_\tau)-\hat{f}_{t+1}(x^\prime_\tau)\right)-\left({f}(x_\tau)-{f}(x^\prime_\tau)\right)\right)^2\nonumber\\
  &\leq\beta(\epsilon, \nicefrac{\delta}{2}, t), 
\end{align}
and 
\begin{equation}
   \tf\in\mathcal{B}_f^{t+1}, 
\end{equation}
where $\beta(\epsilon, \nicefrac{\delta}{2}, t)=\frac{\underline{\sigma^\prime}^2}{H_\sigma}\left(\beta_2(\epsilon,\nicefrac{\delta}{2}, t)+2\beta_1(\epsilon, \nicefrac{\delta}{2}, t)\right)=\mathcal{O}\left(\sqrt{t\log\frac{t\mathcal{N}(\mathcal{B}_f, \epsilon,\|\cdot\|_\infty)}{\delta}}+\epsilon t+\epsilon^2t\right)$, with $\beta_2(\epsilon, \delta, t)=8H_\sigma\bar{\sigma^\prime}^2\epsilon^2t+2C_L\epsilon t+\sqrt{8{tB_p^2\log\frac{\pi^2t^2\mathcal{N}(\mathcal{B}_f,\epsilon, \|\cdot\|_\infty)}{3\delta}}{}}$ and the constants $\underline{\sigma^\prime}, H_\sigma, \bar{\sigma^\prime}, B_p$ as defined in Appendix~\ref{sec:link_prop}. 
\end{lemma}
Lem.~\ref{thm:obj_ell_po} highlights that with high probability, all the functions in the confidence set have difference values over the historical sample points that lie in a ball with the ground-truth function difference value as the center and $\sqrt{\beta(\epsilon, \nicefrac{\delta}{2}, t)}$ as the radius. Lem.~\ref{thm:obj_ell_po} indicates that our likelihood-based learning scheme can gradually learn the function differences $\tf(x_\tau)-\tf(x_\tau^\prime)$ but not the absolute value $\tf(x_\tau)$. This is reasonable since shifting $f$ by a constant will not change the distribution of preference feedback. 

Furthermore, to derive an error bound over a new pair $(x,x^\prime)$, we need to quantify the uncertainty of $\ff(x)-\ff(x^\prime)$, where $\ff\in\mathcal{B}_f$. Since $-\ff\in\mathcal{B}_f$ by the definition of $\mathcal{B}_f$, it can be seen that $\ff(x)-\ff(x^\prime)\in\mathcal{B}_{ff^\prime}$, where 
\begin{equation}
\mathcal{B}_{ff^\prime}\defeq\{F(x,x^\prime)=\ff(x)+\ff^\prime(x^\prime)|\ff,\ff^\prime\in\mathcal{B}_f\}.
\end{equation}
Indeed, $\mathcal{B}_{ff^\prime}$ is the ball with radius $2B$ in the RKHS equipped with the additive kernel function $k^{ff^\prime}((x, x^\prime), (\bar{x}, \bar{x}^\prime))\defeq k(x, \bar{x})+k(x^\prime, \bar{x}^\prime)$, which we term as the augmented RKHS here, and inner product $\langle f_1+f_1^\prime, f_2+f_2^\prime\rangle_{k^{ff^\prime}}\defeq\langle f_1, f_2\rangle_{k}+\langle f_1^\prime, f_2^\prime\rangle_{k}$. The readers are referred to \cite{christmann2012consistency,kandasamy2015high} for more details of the additive kernel and the corresponding RKHS. 
To quantify the uncertainty of a new pair $(x, x^\prime)$, we further introduce the function,
\begin{align}
\label{eqn:ff_sigma_def}
\left(\sigma^{ff^\prime}_{t}(\omega)\right)^2 &={k}^{ff^\prime}\left(\omega, {\omega}\right)\\
-{k}^{ff^\prime}(&\omega_{1:t-1}, \omega)^\top\left({K}^{ff^\prime}_{t-1}+ \lambda I\right)^{-1} {k}^{ff^\prime}\left(\omega_{1:t-1}, {\omega}\right),\nonumber 
\end{align}
where $\omega\defeq(x, x^\prime)$, $\omega_{1:t-1}\defeq((x_\tau, x_\tau^\prime))_{\tau=1}^{t-1}$, $K_{t-1}^{ff^\prime}\defeq(k^{ff^\prime}((x_{\tau_1}, x_{\tau_1}^\prime), (x_{\tau_2}, x_{\tau_2}^\prime)))_{\tau_1\in[t-1], \tau_2\in[t-1]}$, and $\lambda$ is a positive regularization constant. 

\begin{theorem}[\textbf{Duel-wise Error Bound}]
\label{thm:bound_duelwise_error} 
 For any estimate $\hat{f}_{t+1}\in\mathcal{B}_f^{t+1}$ measurable with respect to $\mathcal{F}_{t}$, we have, with probability at least $1-{\delta}$, $\forall t\geq1, (x,x^\prime)\in\mathcal{X}\times\mathcal{X}$,  
    \begin{align}
     &\big|{(\hat{f}_{t+1}(x)-\hat{f}_{t+1}(x^\prime))-(\tf(x)-\tf(x^\prime))}\big|\nonumber\\
     \leq&\;\;2\left(2B+\lambda^{-\nicefrac{1}{2}}\sqrt{\beta(\epsilon, \nicefrac{\delta}{2}, t)}\right)\sigma^{ff^\prime}_{t+1}((x, x^\prime)). 
    \end{align}    
\end{theorem}
\begin{remark}
In preferential BO, we do not get the scalar value of $f(x)-f(x^\prime)$. Hence, $\sigma_t^{ff^\prime}$ can not be interpreted as the posterior standard deviation as in~\cite{srinivas2012information}. However, it turns out that $\sigma_t^{ff^\prime}$, as a measure of uncertainty, still accounts for a factor of the duel-wise error. 
\end{remark}

To characterize the complexity of this augmented RKHS, we use the maximum information gain~\cite{srinivas2012information}, 
\begin{equation} 
\label{eqn:info_gain_def}
\gamma^{ff^\prime}_T\defeq \max_{\Omega \subset {\mathcal{X}\times\mathcal{X}} ;|\Omega|=T} \frac{1}{2} \log \left|I+\lambda^{-1}{K}^{ff^\prime}_{\Omega}\right|,
\end{equation}
where ${K}^{ff^\prime}_{\Omega}=\left(k^{ff^\prime}((x, x^\prime), (\bar{x}, \bar{x}^\prime))\right)_{(x, x^\prime), (\bar{x}, \bar{x}^\prime) \in\Omega}$.

\section{Algorithm}
\subsection{Principled Optimistic Algorithm}
We are now ready to give the optimistic algorithm in Alg.~\ref{alg:opt_pref_BO}. 

\begin{algorithm}[htbp!]
\caption{\textbf{P}rincipled \textbf{O}ptimistic \textbf{P}referential \textbf{B}ayesian \textbf{O}ptimization (\textbf{POP-BO}).}
\label{alg:opt_pref_BO}
\begin{algorithmic}[1]
\normalsize
\STATE Given the initial point $x_0\in\mathcal{X}$ and set $\mathcal{B}^1_f=\mathcal{B}_f$. 
\FOR{$t\in[T]$}
\STATE 
Set the reference point $x_t^\prime=x_{t-1}$.~\alglinelabel{alg_line:gen_ref} 
 \STATE Compute \\
 \vspace{-3ex}$$x_t\in\arg\max_{x\in\mathcal{X}}\max_{\ff\in\mathcal{B}_f^t}(\tilde{f}(x)-\tilde{f}(x_t^\prime)),$$\vspace{-2.8ex}\\ 
 with the inner optimal function denoted as $\tilde{f}_t$.\alglinelabel{alg_line:opt_get_x}
 \STATE Query the comparison oracle to get the feedback result $\mathbf{1}_t$ and append the new data to $\mathcal{D}_t$. 
 \STATE Update the maximum likelihood estimator $\hat{f}^{\mathrm{MLE}}_{t}$ and the posterior confidence set $\mathcal{B}^{t+1}_f$.
\ENDFOR
\end{algorithmic}
\end{algorithm}
The key to Alg.~\ref{alg:opt_pref_BO} is line~\ref{alg_line:opt_get_x}. The idea is to maximize the optimistic advantage of $\ff(x)$ as compared to $\ff(x_t^\prime)$ with the uncertainty of the black-box function $\ff\in\mathcal{B}_f^t$.   


In line~\ref{alg_line:gen_ref}, we set the reference point $x_t^\prime$ as the last generated point $x_{t-1}$. In practice, this may correspond to two possible scenarios. In the first, each comparison requires one experiment, such as image quality comparison. In this case, we only need to set one of the compared pair as the last newly generated solution. 
While in the other scenario, comparing $x_t$ and $x_t^\prime$ needs separate experiments for $x_t$ and $x_t^\prime$. For example, when optimizing the building thermal comfort, the occupants need to experience both thermal conditions to report preference. If at step $t$, the oracle still has memory about the experience with input $x_{t-1}$, we can directly compare $x_t$ and $x_{t-1}$. In this case, setting $x_t^\prime$ to be $x_{t-1}$ saves the experimental expense with $x_t^\prime$.

For online applications, cumulative regret is more of our interest. However, for an offline optimization setting, it may be of more interest to identify one near-optimal solution to report. Unlike in the scalar evaluation setting, where we can directly use the scalar value to report the best observed solution, we can not directly identify the best sampled solution in the preferential Bayesian optimization scenario. To address this issue, we report the solution $x_{t^\star}$, where 
\begin{equation}
t^\star\in\arg\min_{t\in[T]} 2\left(2B+\lambda^{-\nicefrac{1}{2}}\sqrt{\beta(\epsilon, \nicefrac{\delta}{2}, t)}\right)\sigma_t^{ff^\prime}((x_t, x_t^\prime)).\label{eq:best_sample_report}
\end{equation}
The idea is that although the best sample may not be known, we can derive a solution by minimizing the known term $2(2B+\lambda^{-\nicefrac{1}{2}}\sqrt{\beta(\epsilon, \nicefrac{\delta}{2}, t)})\sigma_t^{ff^\prime}((x_t, x_t^\prime))$ to find a solution $x_{t^\star}$ to report. Indeed, this term upper bounds the uncertainty of the optimistic advantage~(as shown in Thm.~\ref{thm:bound_duelwise_error}). Hence, the smaller it is, the more certain that $f(x_t)$ is close to the ground-truth optimal value. At step $t$, we can report the current estimated solution with index $\tau^\star(t)$ satisfying a similar formula to Eq.~\meqref{eq:best_sample_report}.

\subsection{Efficient Computations}
Line~\ref{alg_line:opt_get_x} in Alg.~\ref{alg:opt_pref_BO} requires solving a nested optimization problem with inner variables in an infinite-dimensional function space. The update of the maximum likelihood estimator also requires solving an optimization problem with an infinite-dimensional function as the decision variable. These are in general not tractable in their current forms. Fortunately, we can reduce the infinite-dimensional problems to finite-dimensional ones, thanks to the structures of the problem and the representer theorem~\cite{scholkopf2001generalized}. 

\textbf{Maximum likelihood estimation}.
Since the log-likelihood function 
\begin{align}
\label{eq:ltff}
\ell_t(\ff)=&\log\Prob_{\ff}((x_\tau, x_\tau^\prime,\onetau)_{\tau=1}^t)\\
=&\sum_{\tau=1}^t \left(z_\tau\one_\tau+z_\tau^\prime(1-\one_\tau)\right)-\sum_{\tau=1}^t\log\left(e^{z_\tau}+e^{z_\tau^\prime}\right)\nonumber
\end{align}
only depends on the function value $(z_\tau, z_\tau^\prime)=(\ff(x_\tau), \ff(x_\tau^\prime))$, we only need to optimize over $(z_\tau, z_\tau^\prime)$ subject to that they are functions in $\Hil_k$ with norm less or equal to $B$. 
Furthermore, Alg.~\ref{alg:opt_pref_BO} sets
$
x_\tau^\prime=x_{\tau-1}
$ and thus $z_\tau^\prime=z_{\tau-1}$. So we can reduce the optimization variables to only $(z_\tau)_{\tau=0}^t$. Hence, Eq.~\meqref{eq:ltff} is reduced to the following log-likelihood function that only depends on $(z_\tau)_{\tau=0}^t$,
\begin{align}
   &\ell(Z_{0:t}|\mathcal{D}_{t})\\ 
  \defeq &\; Z_{1:t}^\top \one_{1:t}+{Z_{0:t-1}}^\top(1-\one_{1:t})-\sum_{\tau=1}^t\log\left(e^{z_\tau}+e^{z_{\tau-1}}\right),\nonumber
\end{align}
where $Z_{0:t}\defeq(z_\tau)_{\tau=0}^t$, $Z_{1:t}\defeq(z_\tau)_{\tau=1}^t$, $Z_{0:t-1}\defeq(z_\tau)_{\tau=0}^{t-1}$ and $\one_{1:t}=(\onetau)_{\tau=1}^t$.

By the representer theorem~\cite{scholkopf2001generalized}, the maximum likelihood estimation problem can be solved via,
\begin{equation}
\begin{aligned}
\label{eqn:MLE_prob}
\ell_t(\hat{f}_t^\mathrm{MLE})=\max_{Z_{0:t}\in\mathbb{R}^{t+1}}&\quad \ell(Z_{0:t}|\mathcal{D}_{t})\\
   \textrm{subject to}&\quad 
Z_{0:t}^{\top}K_{0:t}^{-1}
Z_{0:t} \leq B^2, 
\end{aligned}
\end{equation}
where 
$K_{0:t}\defeq(k(x_{\tau_1}, x_{\tau_2}))_{\tau_1\in\{0\}\cup[t], \tau_2\in\{0\}\cup[t]}$. The constraint restricts that the function values need to come from a function inside the function space ball $\mathcal{B}_f$, where the left-hand side is indeed the minimum norm square of the possible interpolant through $\{(x_\tau, z_\tau)\}_{\tau=0}^t$ as shown in~\cite{wendland2004scattered}.  It can be checked that the maximization problem in Eq.~\meqref{eqn:MLE_prob} has a concave objective~(as shown in Appendix~\ref{app:pre}) with a convex feasible set. Thus, the problem in Eq.~\meqref{eqn:MLE_prob} is tractable via convex optimization.    

\textbf{Generating new sample point}. On the line~\ref{alg_line:opt_get_x} of Alg.~\ref{alg:opt_pref_BO}, a bi-level optimization problem needs to be solved, where the inner-level part has an infinite-dimensional function variable. The inner optimization problem has the form,
\begin{equation} 
\begin{aligned}
    \label{eqn:generic_inner_prob}
\max_{\tilde{f}}&\quad \tilde{f}(x)-\tilde{f}(x_t) \\
   \textrm{subject to}&\quad \tilde{f}\in\mathcal{B}_f,\\
   &\quad \ell_t(\tilde{f})\geq \ell_t(\hat{f}^\mathrm{MLE}_t)-\beta_1(\epsilon, \delta, t),
\end{aligned}
\end{equation}
where $\beta_1(\epsilon, \delta, t)$ is as given in Thm.~\ref{thm:conf_set_f}.
Similar to the representer theorem, we have,
\begin{lemma}
\label{thm:inner_finite_red}
Prob.~\meqref{eqn:generic_inner_prob} can be equivalently reduced to,
\begin{equation}    
\begin{aligned}
    \label{eqn:reform_inner_prob_to_fin}
\max_{Z_{0:t}\in\mathbb{R}^{t+1}, z\in\mathbb{R}}&\quad z- z_t \\
   \text{subject to}&\quad \left[\begin{array}{l}
Z_{0:t} \\
z
\end{array}\right]^{\top}K_{0:t,x}^{-1}\left[\begin{array}{l}
Z_{0:t} \\
z
\end{array}\right]   \leq B^2, \\
   &\quad \ell(Z_{0:t}|\mathcal{D}_t)\geq \ell_t(\hat{f}^\mathrm{MLE}_t)-\beta_1(\epsilon, \delta, t),
\end{aligned}
\end{equation}
where 
\begin{equation} 
\begin{aligned}
K_{0:t,x}=\left[\begin{array}{cc}
K_{0:t} & (k(x_\tau, x))_{\tau=0}^t \\
{(k(x_\tau, x))_{\tau=0}^t}^\top & k(x, x) 
\end{array}\right].
\end{aligned}
\end{equation}
\end{lemma}
Similarly, it can be checked that the Prob.~\meqref{eqn:reform_inner_prob_to_fin} is convex. 

For low-dimensional $x$, the outer-level problem can be solved via grid search. For medium-dimensional problems, we can optimize the inner/outer variables using a gradient-based/zero-order optimization method. Alternatively, we can jointly optimize $x, Z_{0:t}$, and $z$ by a nonlinear programming solver from multiple random initial conditions. \rev{That is, we add $x$ as another optimization variable as shown in the Prob.~\meqref{eqn:reform_merge_prob_to_fin},
 \begin{equation}    
\begin{aligned}
    \label{eqn:reform_merge_prob_to_fin}
\max_{Z_{0:t}\in\mathbb{R}^{t+1}, z\in\mathbb{R}, x\in\mathcal{X}}&\quad z- z_t \\
   \text{subject to}&\quad\text{Constraints of Prob.~\meqref{eqn:reform_inner_prob_to_fin}}.
\end{aligned}
\end{equation}   
More details on this joint optimization approach is in Appendix~\ref{app_sec:joint_opt}.
}

\begin{remark}
   We add a matrix $\epsilon_KI$ to $K_{0:t}$ and $K_{0:t, x}$ before inversion to avoid numerical issue, where $\epsilon_K>0$ is small. 
\end{remark}

\rev{
\begin{remark}
    In this paper, we mainly consider the setting where in each step, the preference is queried over two candidate points. Our Alg.~\ref{alg:opt_pref_BO} and the efficient computation schemes in this section can be easily extended to multiple-choice setting, where in each step, the best or most preferred point is queried over a batch of candidates. The detailed discussion is in Appendix~\ref{app_sec:ext_multi_choice}.        
\end{remark}
}

\section{Theoretical Analysis}
We first introduce the performance metrics to use. As in the standard Bayesian optimization setting~(\cite{srinivas2012information}), cumulative regret is used as defined in Eq.~\meqref{eq:cumu_reg_def},
\begin{equation}
   R_T\defeq \sum_{t=1}^T\left(f(x^\star) - f(x_t)\right),\label{eq:cumu_reg_def} 
\end{equation}
where $x^\star\in\arg\max_{x\in\mathcal{X}} f(x)$.

\begin{remark}
    The cumulative regret $R_T$ as defined in Eq.~\meqref{eq:cumu_reg_def} does not explicitly consider the sub-optimality of the reference point $x_t^\prime$. However, since $x_t^\prime=x_{t-1}$, the cumulative regret of the reference points is the same as $R_T$ in Eq.~\meqref{eq:cumu_reg_def}, \rev{up to the difference of the first/last term}.  
\end{remark}

Cumulative regret is of interest in the online setting. In the offline optimization setting, it is of more interest to analyze the sub-optimality of the final reported solution, i.e.,
\begin{equation}
    f(x^\star)-f(x_{t^\star}),
\end{equation}
where ${x}_{t^\star}$ is the final reported solution as defined in Eq.~\meqref{eq:best_sample_report}. 

\subsection{Regret Bound and Convergence Rate}
\begin{theorem}[\textbf{Cumulative Regret Bound}]
\label{thm:regret_bound_general_kernel}
 With probability at least $1-\delta$, the cumulative regret of Alg.~\ref{alg:opt_pref_BO} satisfies,   
   \begin{equation}
     R_T=\mathcal{O}\left(\sqrt{\beta_T\gamma^{ff^\prime}_T T}\right),
   \end{equation}
 where $$\beta_T=\beta(\nicefrac{1}{T}, \delta, T)=\mathcal{O}\left(\sqrt{T\log\frac{T\mathcal{N}(\mathcal{B}_f, \nicefrac{1}{T},\|\cdot\|_\infty)}{\delta}}\right).$$ 
\end{theorem}
\begin{remark}[\textbf{Differentiate from GP-UCB regret}]
Our bound has a similar form as compared to the well-known regret bound for standard GP-UCB type algorithms~\cite{srinivas2012information,chowdhury2017kernelized}. However, the $\beta_T$ term here is significantly different from that in the existing literature~(e.g., in Thm. 3 in~\cite{srinivas2012information}). It is derived specifically for the preferential BO and will lead to a bit larger bound for specific kernels in Sec.~\ref{sec:kern_spec_bound}. 

\end{remark}

\rev{We highlight that Thm.~\ref{thm:regret_bound_general_kernel} provides the \emph{first-of-its-kind} information-theoretic bound on the cumulative regret of preferential BO, which further allows us to derive} a convergence rate for the reported solution $x_{t^\star}$ in Thm.~\ref{thm:solution_report_and_conv_rate}.  
\begin{theorem}[\textbf{Convergence Guarantee}]
\label{thm:solution_report_and_conv_rate}Let $t^\star$ be defined as in Eq.~\meqref{eq:best_sample_report}. With probability at least $1-\delta$,
\begin{equation}
   f(x^\star)-f(x_{t^\star})\leq\mathcal{O}\left(\frac{\sqrt{\beta_T\gamma^{ff^\prime}_T }}{\sqrt{T}}\right).
   \end{equation}
\end{theorem}
Thm~\ref{thm:solution_report_and_conv_rate} highlights that by minimizing the known term $2\big(2B+\lambda^{-\nicefrac{1}{2}}\sqrt{\beta(\epsilon, \frac{\delta}{2}, t)}\big)\sigma_t^{ff^\prime}((x_t, x_t^\prime))$, the reported final solution $x_{t^\star}$ has a guaranteed convergence rate.

\subsection{Kernel-Specific Bounds and Rates}
\label{sec:kern_spec_bound}

In this section, we show kernel-specific bounds for the regret and convergence rate for the reported solution. The explicit forms of the considered kernels are given in Appendix~\ref{app:spec_kerfuncs}.

\begin{theorem}[\textbf{Kernel-Specific Regret Bounds}]
\label{thm:kern_spec_bound}
    Setting $\epsilon=\nicefrac{1}{T}$ and running our POP-BO algorithm in Alg.~\ref{alg:opt_pref_BO},  
    \vspace{-2ex}
   \begin{enumerate}
   \setlength\itemsep{0.02em}
       \item  If $k(x,y)=\langle x, y\rangle$, we have,
   \begin{equation}
      R_T={\mathcal{O}}\left(T^{\nicefrac{3}{4}}(\log T)^{\nicefrac{3}{4}}\right).  
   \end{equation} 
   \item If $k(x,y)$ is a squared exponential kernel, we have,
 \begin{equation}    
       R_T=\mathcal{O}\left(T^{\nicefrac{3}{4}}(\log T)^{\nicefrac{3}{4}(d+1)}\right).   
 \end{equation}
 \item If $k(x,y)$ is a Mat\'ern kernel, we have,
 \begin{equation}    
         R_T=\mathcal{O}\left(T^{\nicefrac{3}{4}}(\log T)^{\nicefrac{3}{4}}T^{\frac{d}{\nu}\left(\frac{1}{4}+\frac{d+1}{4+2(d+1)\nicefrac{d}{\nu}}\right)}\right),
 \end{equation}
 where $\nu$ is the smooth parameter of the Mat\'ern kernel that is assumed to be large enough such that $\nu>\frac{d}{4}(3+d+\sqrt{d^2+14d+17})=\Theta(d^2)$. 
   \end{enumerate} 
   
\end{theorem}

\begin{remark}[\textbf{Comparison to GP-UCB with Scalar Feedback}]
Interestingly, as compared to the kernel-specific bounds in the scalar evaluation-based optimization~(Fig. 1 in~\cite{srinivas2012information}), the regret bound of preferential Bayesian optimization approximately has an additional factor of $T^{\nicefrac{1}{4}}$. This is reasonable since intuitively, scalar evaluation can imply preference, but not vice versa. Therefore, preference feedback contains less information and thus may suffer from higher regret. Fig.~\ref{fig:cumu_sim_r} in Sec.~\ref{sec:exp_sampled_instance} and Fig.~\ref{fig:cumu_reg_loglog} in Appendix~\ref{app_sec:Rloglog} empirically verify our bounds here.
\end{remark}
We then derive the kernel-specific convergence rates for the reported solution $x_{t^\star}$, as shown in Tab.~\ref{tab:kern_spec_bounds} in the Appendix~\ref{app_sec:kern_spec_conv_rate}.



\section{Experimental Results}
In this section, we compare our method to the state-of-the-art preferential BO methods on sampled instances from Gaussian process, standard test functions, and a thermal comfort optimization problem. The comparison outcome is sampled as assumed in Assump.~\ref{assump:pref_model}. We implement our algorithm based on the Gaussian process package GPy~\cite{gpy2014}. The optimization problems for MLE and generating new samples are formulated and solved using CasADi~\cite{andersson2019casadi} and Ipopt~\cite{wachter2006implementation}. We compare our methods to three baseline methods: dueling Thompson sampling~\cite{gonzalez2017preferential}, skew-GP based preferential BO~\cite{pmlr-v202-takeno23b}, and the qEUBO~\cite{astudillo2023qeubo}. The dueling Thompson sampling method~\cite{gonzalez2017preferential} derives the next pair to compare by maximizing the soft-Copeland's score. The skew-GP based method~\cite{pmlr-v202-takeno23b} applies standard BO algorithms conditioned on the Thompson sampling results on the historical sample points that are consistent with the historical preference feedbacks. The qEUBO~\cite{astudillo2023qeubo} method uses the expected utility of the best option as an acquisition function. More experimental details and results on thermal comfort optimization are put in the Appendix~\ref{app_sec:exp_res_detail}.

\subsection{Sampled Instances from Gaussian Process}
\label{sec:exp_sampled_instance}
In this section, we sample the black-box function $f$ from a Gaussian process with the squared exponential kernel as shown in Appendix~\ref{app:spec_kerfuncs}
where the variance parameter is $9.0$ and the lengthscale is $1.0$. We sampled $30$ instances in total.     

\begin{figure}[htbp]
    \centering
    \includegraphics[width=\plotwidth]{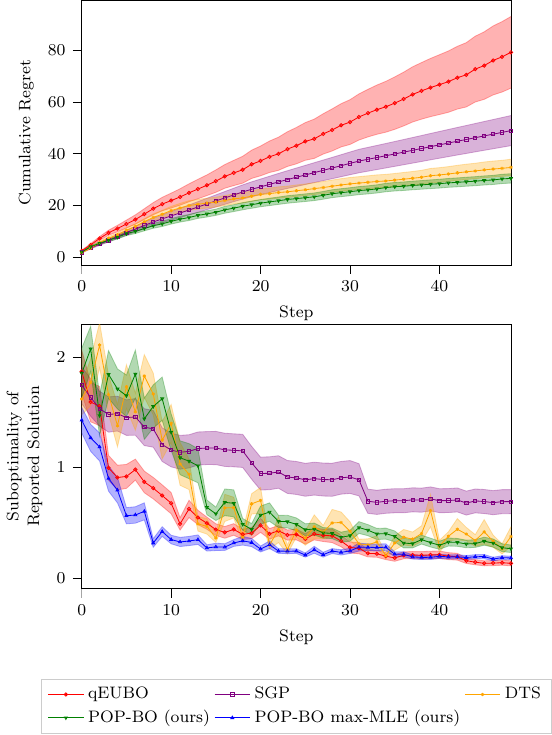}
    \caption{Cumulative regret and the suboptimality of reported solution, where the shaded areas represent $\pm0.1\textsf{ standard deviation}$. qEUBO represents the method in~\cite{astudillo2023qeubo}, {which reports the solution that maximizes the expected objective value conditioned on the historical samples}. SGP represents the skew-GP based method~\cite{pmlr-v202-takeno23b}, {which reports the first point of the duel proposed by the algorithm in the last step}. DTS represents the duelling Thompson sampling method in~\cite{gonzalez2017preferential}, {which reports the Condorcet winner}. 
}
    \label{fig:cumu_sim_r}
\end{figure}

Fig.~\ref{fig:cumu_sim_r} shows the performance comparisons with baselines. Our method achieves the lowest sublinear growth in cumulative regret. It also achieves better/competitive convergence speed for the reported solution as compared to the DTS method, while outperforming the SGP. 

However, our method only uses less than $10\%$ of the computation time as compared to the DTS as shown in Tab.~\ref{tab:comp_time}. The SGP method gets stuck in local optimum because it overly trusts the random preference feedback (hard constraint when doing Thompson sampling). Although the qEUBO method performs slightly better in the reported solution, it suffers from more than $2.5$ times the cumulative regret as compared to ours. Similar to qEUBO~(reporting posterior mean maximizer), we can report the maximizer of the minimum-norm $\hat{f}_t^{\textrm{MLE}}$~(POP-BO max-MLE in Fig.~\ref{fig:cumu_sim_r}) instead of $x_{t^\star}$ in Eq.~\meqref{eq:best_sample_report}, and achieves faster convergence than qEUBO.

\begin{table}[htbp]
    \centering
    \caption{Computation time normalized against the DTS method.}
    \label{tab:comp_time}
    \scalebox{0.87}{{\begin{tabular}{|c|c|c|c|}
      \hline 
       DTS &  qEUBO & SGP & POP-BO~(ours)\\
       \hline 
       $\mathbf{1.0}$ & $0.21$ & $0.07$ & ${0.09}$ \\
       \hline 
    \end{tabular}
    }
    }
\end{table}

\subsection{Test Function Optimization}
In this section, we compare our method to several well-known global optimization test functions~\cite{dixon1978global,molga2005test}, which are divided by the standard deviation of samples over a grid. We run our method multiple times from different random initial points. Tab.~\ref{tab:subopt_testfunc} shows that POP-BO consistently finds better or comparable solutions as compared to other baselines. 

\begin{table}[htbp]
    \centering
    \caption{Suboptimality for the final reported solution after 30 steps. The results (\textsf{mean}\;$\pm$\;\textsf{standard deviation}) are taken over 30 runs with random starting points.}
    \label{tab:subopt_testfunc}
    \scalebox{0.72}{
    {\begin{tabular}{|c|c|c|c|c|}
      \hline 
      Problem &DTS &  qEUBO & SGP & \textbf{POP-BO} (ours)\\
       \hline 
     Beale &  $0.84\pm0.52$ & $0.15\pm0.52$ & $0.10\pm0.19$ & $\mathbf{0.008\pm0.025}$ \\
       \hline
    Branin & $1.35\pm1.16$& $0.71\pm1.16$ & $2.20\pm0.81$ & $\mathbf{0.31\pm0.29}$ \\
    \hline 
     Bukin & \rev{$1.45\pm1.13$} & \rev{$0.59\pm1.20$} & $1.27\pm0.80$ & $\mathbf{0.92\pm0.54}$ \\
    \hline 
   Cross-in-Tray & \rev{$1.56\pm1.39$} & $2.03\pm1.82$ & $1.79\pm1.49$ & $\mathbf{1.38\pm0.97}$ \\
     \hline 
   Eggholder & $3.08\pm0.55$& $3.11\pm0.55$ & $1.87\pm0.94$ & $\mathbf{1.83\pm0.96}$ \\
    \hline 
   Holder Table & $3.21\pm1.38$& $3.20\pm1.38$ & $1.56\pm1.62$ & $\mathbf{1.22\pm1.01}$ \\
    \hline
     Levy13 & $2.36\pm1.22$& $1.06\pm1.22$ & $1.29\pm1.00$ & $\mathbf{0.35\pm0.31}$ \\
    \hline 
    \end{tabular}
    }
    }
\end{table}

\rev{
\subsection{Scalability to Higher Dimension}
To demonstrate the computational scalability of our joint optimization approach~(as shown in Prob.~\meqref{eqn:reform_merge_prob_to_fin}), we consider a set of higher dimensional problems. Due to space limitation, we show the results for the optimization of $12$-dimensional black-box function sampled from a Gaussian process with squared exponential kernel function. More results can be found in Appendix~\ref{app_sec:high_dim_res} and Appendix~\ref{app_subsec:therm_scale_to_higher}. 
The optimization domain is set to be $[0,10]^{12}$. We run $10$ randomly sampled instances for $100$ steps. The average update time per step is only $18.0$ seconds on a personal computer with one \texttt{Intel64 Family 6 Model 142 Stepping 12 GenuineIntel ~1803 Mhz} processor and \texttt{16.0 GB RAM}. This is comparably very small considering that each query to the comparison oracle can be very expensive in practice (e.g., heating the room up to a certain temperature to evaluate occupant comfort, which may take tens of minutes). We compare our method to the SGP baseline, which is one of the state-of-the-art computationally practical preferential Bayesian optimization method. Fig.~\ref{fig:cumu_sim_reg_high_gp_dim12} shows the cumulative regret (in log scale) and the suboptimality of the reported solution for the problem. It can be seen that our algorithm still achieves sublinear regret growth and good convergence for the suboptimality of the reported solution within $100$ steps in this 12-dimensional problem. Fig.~\ref{fig:cumu_sim_reg_high_gp_dim12} \rev{also} shows that our POP-BO has faster convergence speed in higher dimensional problem and thus scales better than the SGP method.       

\begin{figure}[h!]
    \centering   \includegraphics[width=\plotwidth]{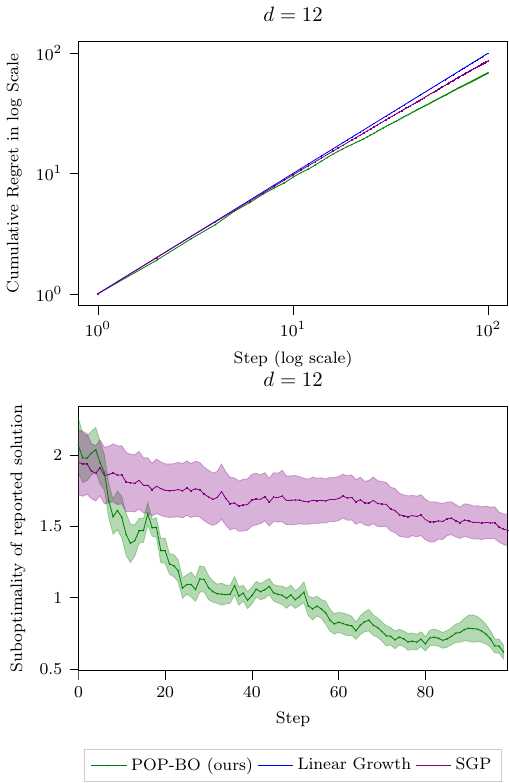}
    \caption{\rev{Cumulative regret in log scale and the suboptimality of reported solution in linear scale for a $12$-dimensional problem sampled from Gaussian process. For reference purpose, we also plot $T$ in the cumulative regret plot in log scale, where the shaded areas represent $\pm0.2\textsf{ standard deviation}$.}   
    }
    \label{fig:cumu_sim_reg_high_gp_dim12}
\end{figure}

}

\section{Conclusion \rev{and Future Work}}
In this paper, we have presented a principled optimistic preferential BO algorithm, based on the likelihood-based confidence set. An efficient computational method is developed to implement the algorithm. We further show an information-theoretic bound on the cumulative regret, a \emph{first-of-its-kind} for preferential BO. We also design a scheme to report an estimated optimal solution, with a guaranteed convergence rate. Experimental results show that our method achieves better or competitive performance as compared to the state-of-the-art heuristics, which, however, do not have theoretical guarantees on regret. Future works include the extension to the safety-critical problem~\cite{berkenkamp2016safe,guo2023safe} and game \rev{theoretical} setting. The likelihood-based confidence set and the error bound in Sec.~\ref{sec:high_conf_set} can also be applied to more scenarios with preference feedback.

\rev{
 \section*{Acknowledgements} 
This research was supported by the Swiss National Science Foundation under NCCR Automation, grant agreement {51NF40\_180545}, the Swiss Federal Office of Energy SFOE as part of the SWEET consortium SWICE, and in part by the Swiss Data Science Center, grant agreement C20-13.
}

\section*{Impact Statement}
This paper presents work whose goal is to advance the field of Machine Learning. There are many potential societal consequences of our work, none which we feel must be specifically highlighted here.

\bibliography{example_paper}
\bibliographystyle{icml2024}

\newpage
\appendix
\onecolumn
\typeout{Single Columnwidth: \the\columnwidth}
Without further notice, all the results shown in this appendix are under the assumptions~\ref{assump:support_set}, \ref{assump:bounded_norm}, \ref{assump:kernel_bound}, and \ref{assump:pref_model}. 
\section{Preliminaries}
\label{app:pre}
To prepare for the proofs of the main results shown in this paper, we first state several useful lemmas.
\begin{lemma}
    \label{lem:}
    The function $\psi(y, y^\prime)=\log(e^y+e^{y^\prime})$ is convex in $(y, y^\prime)$. 
\end{lemma}
\begin{proof}
We calculate the Hessian of the function $\psi$ and derive 
\begin{align}
\label{eqn:reform_inner_prob}
\nabla^2\psi=\frac{e^{y+y^\prime}}{(e^y+e^{y^\prime})^2}\left[\begin{array}{cccc}
1 &-1  \\
 -1 & 1\\
\end{array}\right]\succcurlyeq0. 
\end{align}
Hence, $\psi$ is convex.
    
\end{proof}
Therefore, we can see $\ell(Z_{0:t}|\mathcal{D}_t)$ is concave in $Z_{0:t}$.

\begin{lemma}
    $\forall\ff\in\mathcal{B}_f, x\in\mathcal{X}, \ff(x)\in[-B, B].$
\end{lemma}
\begin{proof}
   $|\ff(x)|=|\langle\ff, k(x,\cdot)\rangle|\leq\norm{\ff}\norm{k(x, \cdot)}\leq B\sqrt{k(x, x)}\leq B$, where the first inequality follows by Cauchy–Schwarz inequality, the second inequality follows by Assump.~\ref{assump:bounded_norm}, and the last inequality follows by Assump.~\ref{assump:kernel_bound}.  
\end{proof}

\section{Properties of the function $\sigma(\cdot)$}
\label{sec:link_prop}
When applying the function $\sigma$ to the difference of objective function $\ff(x)-\ff(x^\prime),\forall\ff\in\mathcal{B}_f$, we have the calculations by single variable calculus,
\begin{align*}
u&\defeq \ff(x)-\ff(x^\prime)\in[-2B, 2B],\\
\sigma(u)&\in[\underline{\sigma}, \bar{\sigma}],\\
\sigma^\prime(u)&=\frac{1}{2+e^u+e^{-u}}\in[\underline{\sigma^\prime}, \bar{\sigma^\prime}],\\
\end{align*}
where $\underline{\sigma}=\nicefrac{1}{(1+e^{2B})}, \bar{\sigma}=\nicefrac{1}{(1+e^{-2B})}$ and $\underline{\sigma^\prime}=\nicefrac{1}{(2+e^{2B}+e^{-2B})}, \bar{\sigma^\prime}=\nicefrac{1}{4}$. We also introduce some constants $B_p=\frac{\bar{\sigma}}{\underline{\sigma}}-\frac{\underline{\sigma}}{\bar{\sigma}}$, $H_\sigma=\frac{1}{2\bar{\sigma}^2}$ and $C_L=1+\frac{{2}}{1+e^{-2B}}$, which will be used in the proof.

\section{Proof of Thm.~\ref{thm:conf_set_f}}
To prepare for the proof of the theorem, we first prove several lemmas.

\begin{lemma} For any fixed $\hat{f}\in\mathcal{B}_f$, we have,
\begin{equation}
  \Prob\left(\log\Prob_{\hat{f}}((x_\tau, x_\tau^\prime,\onetau)_{\tau=1}^t)-\log\Prob_{{f}}((x_\tau, x_\tau^\prime,\onetau)_{\tau=1}^t)\leq \sqrt{32tB^2\log\frac{1}{\delta_t}}\right)\geq 1-\delta_t, 
\end{equation}
where $f$ is the ground-truth function.
\end{lemma}
\begin{proof}
We use $y_\tau$~($y_\tau^\prime$ resp.) to denote $f(x_\tau)$~($f(x_\tau^\prime)$ resp.). We use $z_\tau$~($z_\tau^\prime$ resp.) to denote $\hat{f}(x_\tau)$~($\hat{f}(x_\tau^\prime)$ resp.). And we use $p_\tau$ to denote $\sigma(y_\tau-y_\tau^\prime)$. 
\begin{align*}
&\Prob\left(\log\Prob_{\hat{f}}((x_\tau, x_\tau^\prime,\onetau)_{\tau=1}^t)-\log\Prob_{{f}}((x_\tau, x_\tau^\prime,\onetau)_{\tau=1}^t)\leq \xi\right)\\
=& \Prob\left(\sum_{\tau=1}^t \left((z_\tau-y_\tau)\one_\tau+(z_\tau^\prime-y_\tau^\prime)(1-\one_\tau)\right)-\sum_{\tau=1}^t\log\left(e^{z_\tau}+e^{z_\tau^\prime}\right)+\sum_{\tau=1}^t\log\left(e^{y_\tau}+e^{y_\tau^\prime}\right)\leq \xi\right) \\
=&\Prob\left(\sum_{\tau=1}^t \left((z_\tau-y_\tau)\one_\tau+(z_\tau^\prime-y_\tau^\prime)(1-\one_\tau)\right)-\sum_{\tau=1}^t \left((z_\tau-y_\tau)p_\tau+(z_\tau^\prime-y_\tau^\prime)(1-p_\tau)\right) \leq \xi^\prime\right) 
\end{align*}
where $\xi^\prime=\xi+\sum_{\tau=1}^t\log\left(e^{z_\tau}+e^{z_\tau^\prime}\right)-\sum_{\tau=1}^t\log\left(e^{y_\tau}+e^{y_\tau^\prime}\right)-\sum_{\tau=1}^t \left((z_\tau-y_\tau)p_\tau+(z_\tau^\prime-y_\tau^\prime)(1-p_\tau)\right)$, and the probability $\Prob$ is taken with respect to the randomness from the comparison oracle and the randomness from the algorithm.

It can be checked that $\psi_\tau(y,y^\prime)\defeq\log\left(e^{y}+e^{y^\prime}\right)-p_\tau y-(1-p_\tau)y^\prime$ is a convex function and $\nabla\psi_\tau(y_\tau, y_\tau^\prime)=0$. This implies that $(y_\tau,y_\tau^\prime)$ achieves the minimum for the convex function $\psi_\tau$. Therefore, $$\log\left(e^{y_\tau}+e^{y_\tau^\prime}\right)-p_\tau y_\tau-(1-p_\tau)y_\tau^\prime\leq \log\left(e^{z_\tau}+e^{z_\tau^\prime}\right)-p_\tau z_\tau-(1-p_\tau)z_\tau^\prime.$$ Rearrangement gives,
$$\log\left(e^{z_\tau}+e^{z_\tau^\prime}\right)-\log\left(e^{y_\tau}+e^{y_\tau^\prime}\right)-\left((z_\tau-y_\tau)p_\tau+(z_\tau^\prime-y_\tau^\prime)(1-p_\tau)\right)\geq0.$$
Hence, $\xi^\prime\geq\xi$. Therefore,
\begin{align*}
&\Prob\left(\log\Prob_{\hat{f}}((x_\tau, x_\tau^\prime,\onetau)_{\tau=1}^t)-\log\Prob_{{f}}((x_\tau, x_\tau^\prime,\onetau)_{\tau=1}^t)\leq \xi\right)\\
=&\Prob\left(\sum_{\tau=1}^t \left((z_\tau-y_\tau)\one_\tau+(z_\tau^\prime-y_\tau^\prime)(1-\one_\tau)\right)-\sum_{\tau=1}^t \left((z_\tau-y_\tau)p_\tau+(z_\tau^\prime-y_\tau^\prime)(1-p_\tau)\right) \leq \xi^\prime\right)\\
\geq & \Prob\left(\sum_{\tau=1}^t \left((z_\tau-y_\tau)\one_\tau+(z_\tau^\prime-y_\tau^\prime)(1-\one_\tau)\right)-\sum_{\tau=1}^t \left((z_\tau-y_\tau)p_\tau+(z_\tau^\prime-y_\tau^\prime)(1-p_\tau)\right) \leq \xi\right)\\
\end{align*}

We further notice that 
\begin{equation}
   \E[\left((z_\tau-y_\tau)\one_\tau+(z_\tau^\prime-y_\tau^\prime)(1-\one_\tau)\right)- \left((z_\tau-y_\tau)p_\tau+(z_\tau^\prime-y_\tau^\prime)(1-p_\tau)\right)|\mathcal{F}_{\tau-1}]=0,
\end{equation}
and with probability one, 
\begin{equation}
   \left\lvert{\left((z_\tau-y_\tau)\one_\tau+(z_\tau^\prime-y_\tau^\prime)(1-\one_\tau)\right)- \left((z_\tau-y_\tau)p_\tau+(z_\tau^\prime-y_\tau^\prime)(1-p_\tau)\right)}\right\rvert=\left\lvert{(z_\tau-y_\tau-z_\tau^\prime+y_\tau^\prime)(\onetau-p_\tau)}\right\rvert\leq4B.
\end{equation}

We can thus apply the Azuma-Hoeffding inequality~(see, e.g.,~\cite{lalley2013concentration}).
By Azuma–Hoeffding inequality, 
\begin{align*}
&\Prob\left(\log\Prob_{\hat{f}}((x_\tau, x_\tau^\prime,\onetau)_{\tau=1}^t)-\log\Prob_{{f}}((x_\tau, x_\tau^\prime,\onetau)_{\tau=1}^t)\leq \xi\right)\\
\geq&\Prob\left(\sum_{\tau=1}^t \left((z_\tau-y_\tau)\one_\tau+(z_\tau^\prime-y_\tau^\prime)(1-\one_\tau)\right)-\sum_{\tau=1}^t \left((z_\tau-y_\tau)p_\tau+(z_\tau^\prime-y_\tau^\prime)(1-p_\tau)\right) \leq \xi\right)\\
\geq & 1-\exp\left\{-\frac{\xi^2}{32tB^2}\right\}.
\end{align*}
Set $\exp\left\{-\frac{\xi^2}{32tB^2}\right\}=\delta_t$. 
That is, $\xi=\sqrt{32tB^2\log\frac{1}{\delta_t}}$. We then get the desired result. 
\end{proof}

We then have the following high probability confidence set lemma.
\begin{lemma}
\label{lem:fixed_f_log_P}
For any fixed $\hat{f}\in\mathcal{B}_f$ that is independent of $((x_\tau, x_\tau^\prime,\onetau)_{\tau=1}^t)$, we have, with probability at least $1-\delta$, 
\begin{equation}
  \log\Prob_{\hat{f}}((x_\tau, x_\tau^\prime,\onetau)_{\tau=1}^t)-\log\Prob_{{f}}((x_\tau, x_\tau^\prime,\onetau)_{\tau=1}^t)\leq \sqrt{32tB^2\log\frac{\pi^2t^2}{6\delta}},\;\forall t\geq1. 
\end{equation}
\end{lemma}
\begin{proof}
We use $\mathcal{E}_t$ to denote the event $\log\Prob_{\hat{f}}((x_\tau, x_\tau^\prime,\onetau)_{\tau=1}^t)-\log\Prob_{{f}}((x_\tau, x_\tau^\prime,\onetau)_{\tau=1}^t)\leq \sqrt{32tB^2\log\frac{1}{\delta_t}}$. We pick $\delta_t=\nicefrac{(6\delta)}{(\pi^2 t^2)}$ and have,
\begin{align*}
  &\Prob\left(\log\Prob_{\hat{f}}((x_\tau, x_\tau^\prime,\onetau)_{\tau=1}^t)-\log\Prob_{{f}}((x_\tau, x_\tau^\prime,\onetau)_{\tau=1}^t)\leq \sqrt{32tB^2\log\frac{1}{\delta_t}}, \forall t\geq1\right) \\
  =& 1-\Prob\left(\overline{\cap_{t=1}^\infty \mathcal{E}_t}\right)\\
  =& 1-\Prob\left({\cup_{t=1}^\infty \overline{\mathcal{E}_t}}\right)\\
  \geq& 1-\sum_{t=1}^\infty\Prob\left( \overline{\mathcal{E}_t}\right)\\
  =& 1-\sum_{t=1}^\infty\left(1-\Prob\left({\mathcal{E}_t}\right)\right)\\
 =& 1-\sum_{t=1}^\infty\left(1-\Prob\left({\log\Prob_{\hat{f}}((x_\tau, x_\tau^\prime,\onetau)_{\tau=1}^t)-\log\Prob_{{f}}((x_\tau, x_\tau^\prime,\onetau)_{\tau=1}^t)\leq \sqrt{32tB^2\log\frac{1}{\delta_t}}}\right)\right)\\
 \geq&1-\sum_{t=1}^\infty\delta_t\\
 = &1-\frac{6\delta}{\pi^2}\sum_{t=1}^\infty\frac{1}{t^2}\\
= & 1-\delta. 
\end{align*}
\end{proof}

We then have a lemma to bound the difference of log likelihood when two functions are close in infinity-norm sense. 
\begin{lemma}
\label{lem:close_f}
There exists an independent constant $C_L>0$, such that, $\forall \epsilon>0$, $\forall f_1, f_2\in\mathcal{B}_f$ that satisfies $\|f_1-f_2\|_\infty\leq\epsilon$, we have,
\begin{equation}
     \log\Prob_{{f}_1}((x_\tau, x_\tau^\prime,\onetau)_{\tau=1}^t)-\log\Prob_{{f}_2}((x_\tau, x_\tau^\prime,\onetau)_{\tau=1}^t)\leq C_L\epsilon t. 
\end{equation}
\end{lemma}
\begin{proof}
We use $z_{i,\tau}$~($z_{i,\tau}^\prime$, resp.) to denote $f_i(x_\tau)$~($f_i(x_\tau^\prime)$, resp.), $\forall i\in\{0, 1\}$. 
\begin{align}
 &\log\Prob_{{f}_1}((x_\tau, x_\tau^\prime,\onetau)_{\tau=1}^t)-\log\Prob_{{f}_2}((x_\tau, x_\tau^\prime,\onetau)_{\tau=1}^t)\nonumber\\
=& \sum_{\tau=1}^t \left((z_{1,\tau}-z_{2,\tau})\onetau+(z_{1,\tau}^\prime-z_{2,\tau}^\prime)(1-\onetau)\right)-\sum_{\tau=1}^t\log\left(e^{z_{1,\tau}}+e^{z_{1,\tau}^\prime}\right)+\sum_{\tau=1}^t\log\left(e^{z_{2,\tau}}+e^{z_{2,\tau}^\prime}\right)\label{eq:def_ll}\\
\leq&\epsilon t+\sum_{\tau=1}^t\max_{z,z^\prime\in[-B, B]}\left\|\nabla_{z,z^\prime}\log\left(e^{z}+e^{z^\prime}\right)\right\|\left\|(z_{1,\tau},z_{1,\tau}^\prime)-(z_{2,\tau},z_{2,\tau}^\prime)\right\|\label{inq:assump}\\
\leq&\epsilon t+\sum_{\tau=1}^t\frac{\sqrt{2}}{1+e^{-2B}}\sqrt{2}\epsilon\\
=&\left(1+\frac{{2}}{1+e^{-2B}}\right)\epsilon t,
\end{align}
where the equality~\meqref{eq:def_ll} follows by the definition of log-likelihood function, and the inequality~\meqref{inq:assump} follows by the assumption and the mean-value theorem.   
The conclusion follows by setting $C_L=1+\frac{{2}}{1+e^{-2B}}$.
    
\end{proof}

\emph{\textbf{Main proof}}: We use $\mathcal{N}(\mathcal{B}_f,\epsilon,\|\cdot\|_\infty)$ to denote the covering number of the set $\mathcal{B}_f$, with $(f^\epsilon_i)_{i=1}^{\mathcal{N}(\mathcal{B}_f,\epsilon,\|\cdot\|_\infty)}$ be a set of $\epsilon$-covering for the set $\mathcal{B}_f$. Reset the `$\delta$' in Lem.~\ref{lem:fixed_f_log_P} as $\nicefrac{\delta}{\mathcal{N}(\mathcal{B}_f,\epsilon,\|\cdot\|_\infty)}$ and applying the probability union bound, we have, with probability at least $1-\delta$, $\forall f_i^\epsilon, t\geq1$, 
\begin{equation}
  \log\Prob_{{f}_i^\epsilon}((x_\tau, x_\tau^\prime,\onetau)_{\tau=1}^t)-\log\Prob_{{f}}((x_\tau, x_\tau^\prime,\onetau)_{\tau=1}^t)\leq \sqrt{32tB^2\log\frac{\pi^2t^2\mathcal{N}(\mathcal{B}_f,\epsilon,\|\cdot\|_\infty)}{6\delta}}. \label{inq:covering_num_involved} 
\end{equation}
By the definition of $\epsilon$-covering, there exists $j\in[\mathcal{N}(\mathcal{B}_f,\epsilon,\|\cdot\|_\infty)]$, such that,
\begin{equation}
   \|\hat{f}_{t}^\mathrm{MLE}-f_j^\epsilon\|_\infty\leq\epsilon. 
\end{equation}
Hence, with probability at least $1-\delta$, 
\begin{align*}
  &\log\Prob_{\hat{f}_{t}^\mathrm{MLE}}((x_\tau, x_\tau^\prime,\onetau)_{\tau=1}^t)-\log\Prob_{{f}}((x_\tau, x_\tau^\prime,\onetau)_{\tau=1}^t)\\
  =&\log\Prob_{\hat{f}_{t}^\mathrm{MLE}}((x_\tau, x_\tau^\prime,\onetau)_{\tau=1}^t)-\log\Prob_{{f}_{j}^\epsilon}((x_\tau, x_\tau^\prime,\onetau)_{\tau=1}^t)+\log\Prob_{{f}_{j}^\epsilon}((x_\tau, x_\tau^\prime,\onetau)_{\tau=1}^t)-\log\Prob_{{f}}((x_\tau, x_\tau^\prime,\onetau)_{\tau=1}^t)\\
  \leq& C_L\epsilon t+\sqrt{32tB^2\log\frac{\pi^2t^2\mathcal{N}(\mathcal{B}_f,\epsilon,\|\cdot\|_\infty)}{6\delta}}, 
\end{align*}
where the inequality follows by Lem.~\ref{lem:close_f} and the inequality~\meqref{inq:covering_num_involved}.

\section{Proof of Lem.~\ref{thm:obj_ell_po}}
We first have a lemma.
\begin{lemma}
    We have, 
\begin{equation}
\log \hat{p}-\log p \leq\frac{1}{p}(\hat{p}-p)-H_\sigma(\hat{p}-p)^2, \forall p,\hat{p}\in[\underline{\sigma}, \bar{\sigma}], 
\end{equation}
where $H_\sigma=\frac{1}{2\bar{\sigma}^2}$.
\end{lemma}
\begin{proof}
   Let $\zeta(\hat{p})=\log \hat{p}- \log p -\frac{1}{p}(\hat{p}-p)+H_\sigma(\hat{p}-p)^2, \forall p,\hat{p}\in[\underline{\sigma}, \bar{\sigma}].$ We have,
   $$
   \zeta^\prime(\hat{p})=\frac{1}{\hat{p}}-\frac{1}{p}+2H_\sigma(\hat{p}-p)=(\hat{p}-p)\left(\frac{1}{\bar{\sigma}^2}-\frac{1}{\hat{p}p}\right), \forall \hat{p}\in[\underline{\sigma}, \bar{\sigma}]. 
   $$
  Since $\forall p,\hat{p}\in[\underline{\sigma}, \bar{\sigma}]$, we have $\frac{1}{\bar{\sigma}^2}-\frac{1}{\hat{p}p}\leq0$. Hence, $\zeta^\prime(\hat{p})\geq0, \forall\hat{p}\in[\underline{\sigma}, p]$ and $\zeta^\prime(\hat{p})\leq0, \forall\hat{p}\in[p, \bar{\sigma}]$. Therefore, $\zeta(\hat{p})$ achieves the maximum over $[\underline{\sigma}, \bar{\sigma}]$ at the point $p$. So $\zeta(\hat{p})\leq\zeta(p)=0$. Rearrangement then gives the desired result.  
\end{proof}

For any fixed function $\hat{f}\in\mathcal{B}_f$, we use the notations $\hat{p}_\tau=\sigma(\hat{f}(x_\tau)-\hat{f}(x_\tau^\prime))\in[\underline{\sigma}, \bar{\sigma}]$ and ${p}_\tau=\sigma({f}(x_\tau)-{f}(x_\tau^\prime))\in[\underline{\sigma}, \bar{\sigma}]$. We have,
\begin{align*}
&\log{\Prob_{\hat{f}}}((x_\tau, x_\tau^\prime,\onetau)_{\tau=1}^t)-\log{\Prob_{{f}}}((x_\tau, x_\tau^\prime,\onetau)_{\tau=1}^t)\\
=&\sum_{\tau=1}^t\left(\log{p_{\hat{f}}}(x_\tau, x_\tau^\prime,\onetau)-\log{p_{{f}}}(x_\tau, x_\tau^\prime,\onetau)\right)\\
=&\sum_{\tau=1}^t\left(\onetau\left(\log{\hat{p}_\tau}-\log{p}_\tau\right)+(1-\onetau)\left(\log{(1-\hat{p}_\tau)}-\log(1-{p}_\tau)\right)\right).
\end{align*}

Hence,
\begin{align*}
&\log{\Prob_{\hat{f}}}((x_\tau, x_\tau^\prime,\onetau)_{\tau=1}^t)-\log{\Prob_{{f}}}((x_\tau, x_\tau^\prime,\onetau)_{\tau=1}^t)\\
=&\sum_{\tau=1}^t\left(\onetau\left(\log{\hat{p}_\tau}-\log{p}_\tau\right)+(1-\onetau)\left(\log{(1-\hat{p}_\tau)}-\log(1-{p}_\tau)\right)\right)\\
\leq&\sum_{\tau=1}^t\left(\onetau\left(\frac{\hat{p}_\tau-p_\tau}{p_\tau}-H_\sigma\left(\hat{p}_\tau-p_\tau\right)^2\right)+(1-\onetau)\left(\frac{{p}_\tau-\hat{p}_\tau}{1-p_\tau}-H_\sigma\left(\hat{p}_\tau-p_\tau\right)^2\right)\right)
\end{align*}
Rearrangement gives,
\begin{equation}
H_\sigma\sum_{\tau=1}^t\left(\hat{p}_\tau-p_\tau\right)^2+\log{\Prob_{\hat{f}}}((x_\tau, x_\tau^\prime,\onetau)_{\tau=1}^t)-\log{\Prob_{{f}}}((x_\tau, x_\tau^\prime,\onetau)_{\tau=1}^t)
\leq\sum_{\tau=1}^t\left(\onetau\frac{\hat{p}_\tau-p_\tau}{p_\tau}+(1-\onetau)\frac{{p}_\tau-\hat{p}_\tau}{1-p_\tau}\right). \label{inq:bound_pdiff}
\end{equation}
We then have the following lemma,
\begin{lemma}
For any fixed $\hat{f}\in\mathcal{B}_f$ and $\forall t\geq1$, we have, with probability at least $1-\delta_t$, 
 \begin{equation}
 \Prob\left(H_\sigma\sum_{\tau=1}^t\left(\hat{p}_\tau-p_\tau\right)^2\leq\log{\Prob_{{f}}}((x_\tau, x_\tau^\prime,\onetau)_{\tau=1}^t)-\log{\Prob_{\hat{f}}}((x_\tau, x_\tau^\prime,\onetau)_{\tau=1}^t)+\sqrt{2{tB_p^2\log\frac{1}{\delta_t}}{}}\right)\geq 1-\delta_t.    
 \end{equation}

\end{lemma}
\begin{proof}
Since $\E\left[\onetau\frac{\hat{p}_\tau-p_\tau}{p_\tau}+(1-\onetau)\frac{{p}_\tau-\hat{p}_\tau}{1-p_\tau}|\mathcal{F}_{\tau-1}\right]=\E\left[p_\tau\frac{\hat{p}_\tau-p_\tau}{p_\tau}+(1-p_\tau)\frac{{p}_\tau-\hat{p}_\tau}{1-p_\tau}|\mathcal{F}_{\tau-1}\right]=0$ and with probability one, 
\begin{align}
\left\lvert{\onetau\frac{\hat{p}_\tau-p_\tau}{p_\tau}+(1-\onetau)\frac{{p}_\tau-\hat{p}_\tau}{1-p_\tau}}\right\rvert&\leq\onetau\left\lvert{\frac{\hat{p}_\tau-p_\tau}{p_\tau}}\right\rvert+(1-\onetau)\left\lvert{\frac{{p}_\tau-\hat{p}_\tau}{1-p_\tau}}\right\rvert\\
&=\onetau\left\lvert{\frac{\hat{p}_\tau}{p_\tau}-1}\right\rvert+(1-\onetau)\left\lvert{\frac{1-\hat{p}_\tau}{1-p_\tau}-1}\right\rvert\\
&\leq \frac{\bar{\sigma}}{\underline{\sigma}}-\frac{\underline{\sigma}}{\bar{\sigma}}=B_p,
\end{align}
where the last inequality follows by that $\hat{p}_\tau, p_\tau, 1-\hat{p}_\tau, 1-p_\tau\in[\underline{\sigma}, \bar{\sigma}]$.
Thus we can apply the Azuma–Hoeffding inequality. By Azuma–Hoeffding inequality, we have,
\begin{equation}
\Prob\left(\sum_{\tau=1}^t\left(\onetau\frac{\hat{p}_\tau-p_\tau}{p_\tau}+(1-\onetau)\frac{{p}_\tau-\hat{p}_\tau}{1-p_\tau}\right)\leq\xi\right)\geq 1-\exp\left\{-\frac{\xi^2}{2tB_p^2}\right\}. 
\end{equation}
We set $\exp\left\{-\frac{\xi^2}{2tB_p^2}\right\}=\delta_t$, and derive
\begin{equation}
 \Prob\left(\sum_{\tau=1}^t\left(\onetau\frac{\hat{p}_\tau-p_\tau}{p_\tau}+(1-\onetau)\frac{{p}_\tau-\hat{p}_\tau}{1-p_\tau}\right)\leq\sqrt{{2tB_p^2\log\frac{1}{\delta_t}}{}}\right)\geq 1-\delta_t. \label{inq:Pbound} 
\end{equation}
Combining the inequality~\meqref{inq:bound_pdiff} and the inequality~\meqref{inq:Pbound}, the desired result is derived.   
    
\end{proof}

\begin{lemma}
\label{lem:p_square_bound}
For any fixed $\hat{f}\in\mathcal{B}_f$, we have, with probability at least $1-\delta$,
\begin{equation}
  H_\sigma\sum_{\tau=1}^t\left(\hat{p}_\tau-p_\tau\right)^2\leq\log{\Prob_{{f}}}((x_\tau, x_\tau^\prime,\onetau)_{\tau=1}^t)-\log{\Prob_{\hat{f}}}((x_\tau, x_\tau^\prime,\onetau)_{\tau=1}^t)+\sqrt{{2tB_p^2\log\frac{\pi^2t^2}{6\delta}}},\;\;\forall t\geq1. 
\label{inq:p_square_bound}
\end{equation}
\end{lemma}
\begin{proof}
We use $\mathcal{E}_t$~\footnote{With abuse of notation here. $\mathcal{E}_t$ is only a local notation in this proof here.} to denote the event $H_\sigma\sum_{\tau=1}^t\left(\hat{p}_\tau-p_\tau\right)^2\leq\log{\Prob_{{f}}}((x_\tau, x_\tau^\prime,\onetau)_{\tau=1}^t)-\log{\Prob_{\hat{f}}}((x_\tau, x_\tau^\prime,\onetau)_{\tau=1}^t)+\sqrt{{2tB_p^2\log\frac{1}{\delta_t}}{}}$ and pick $\delta_t=\nicefrac{(6\delta)}{(\pi^2 t^2)}$.
We have,
\begin{align*}
  &\Prob\left(H_\sigma\sum_{\tau=1}^t\left(\hat{p}_\tau-p_\tau\right)^2\leq\log{\Prob_{{f}}}((x_\tau, x_\tau^\prime,\onetau)_{\tau=1}^t)-\log{\Prob_{\hat{f}}}((x_\tau, x_\tau^\prime,\onetau)_{\tau=1}^t)+\sqrt{{2tB_p^2\log\frac{1}{\delta_t}}{}}, \forall t\geq1\right) \\
  =& 1-\Prob\left(\overline{\cap_{t=1}^\infty \mathcal{E}_t}\right)\\
  =& 1-\Prob\left({\cup_{t=1}^\infty \overline{\mathcal{E}_t}}\right)\\
  \geq& 1-\sum_{t=1}^\infty\Prob\left( \overline{\mathcal{E}_t}\right)\\
  =& 1-\sum_{t=1}^\infty\left(1-\Prob\left({\mathcal{E}_t}\right)\right)\\
 =& 1-\sum_{t=1}^\infty\left(1-\Prob\left(H_\sigma\sum_{\tau=1}^t\left(\hat{p}_\tau-p_\tau\right)^2\leq\log{\Prob_{{f}}}((x_\tau, x_\tau^\prime,\onetau)_{\tau=1}^t)-\log{\Prob_{\hat{f}}}((x_\tau, x_\tau^\prime,\onetau)_{\tau=1}^t)+\sqrt{{2tB_p^2\log\frac{1}{\delta_t}}{}}\right)\right)\\
 \geq&1-\sum_{t=1}^\infty\delta_t\\
 = &1-\frac{6\delta}{\pi^2}\sum_{t=1}^\infty\frac{1}{t^2}\\
= & 1-\delta. 
\end{align*}
\end{proof}

\textbf{\emph{Main Proof}}: Resetting the `$\delta$' in Lem.~\ref{lem:p_square_bound} to be $\nicefrac{\delta}{\mathcal{N}(\mathcal{B}_f, \epsilon, \|\cdot\|_\infty)}$, we can guarantee the Eq.~\meqref{inq:p_square_bound} holds for all the function in an $\epsilon$-covering of $\mathcal{B}_f$ with probability at least $1-\delta$, by applying the probability union bound.   

For any $\hat{f}_{t+1}\in\mathcal{B}_f^{t+1}\subset\mathcal{B}_f$, there exists a function in the $\epsilon$-covering of $\mathcal{B}_f$, which we set to be $\hat{f}$, such that $\|\hat{f}_{t+1}-\hat{f}\|_\infty\leq\epsilon$. We also use $\hat{p}_\tau^{t+1}$ to denote $\sigma(\hat{f}_{t+1}(x_\tau)-\hat{f}_{t+1}(x_\tau^\prime))$. Thus, we have,
\begin{align}
  &  H_\sigma\sum_{\tau=1}^t\left(\hat{p}^{t+1}_\tau-p_\tau\right)^2\\
  \leq&  2H_\sigma\sum_{\tau=1}^t\left(\hat{p}^{t+1}_\tau-\hat{p}_\tau\right)^2+2H_\sigma\sum_{\tau=1}^t\left(\hat{p}_\tau-{p}_\tau\right)^2 \label{inq:squ1} \\
  \leq& 2H_\sigma\bar{\sigma^\prime}^2\sum_{\tau=1}^t\left((\hat{f}_{t+1}(x_\tau)-\hat{f}_{t+1}(x^\prime_\tau))-(\hat{f}(x_\tau)-\hat{f}(x^\prime_\tau))\right)^2+2H_\sigma\sum_{\tau=1}^t\left(\hat{p}_\tau-{p}_\tau\right)^2\\
  \leq& 8H_\sigma\bar{\sigma^\prime}^2\sum_{\tau=1}^t\epsilon^2+2H_\sigma\sum_{\tau=1}^t\left(\hat{p}_\tau-{p}_\tau\right)^2\label{inq:squ2}\\
  \leq&8H_\sigma\bar{\sigma^\prime}^2\epsilon^2t+\sqrt{8{tB_p^2\log\frac{\pi^2t^2\mathcal{N}(\mathcal{B}_f,\epsilon, \|\cdot\|_\infty)}{6\delta}}{}}+2\left(
 \log{\Prob_{{f}}}((x_\tau, x_\tau^\prime,\onetau)_{\tau=1}^t)-\log{\Prob_{\hat{f}}}((x_\tau, x_\tau^\prime,\onetau)_{\tau=1}^t)\right)\label{inq:squ3}\\
 \leq&C(\epsilon, \delta, t)+2\left(\log{\Prob_{\hat{f}_{t}^\mathrm{MLE}}}((x_\tau, x_\tau^\prime,\onetau)_{\tau=1}^t)-\log{\Prob_{\hat{f}_{t+1}}}((x_\tau, x_\tau^\prime,\onetau)_{\tau=1}^t)\right)\label{inq:squ4}\\
 &+2\left(\log{\Prob_{\hat{f}_{t+1}}}((x_\tau, x_\tau^\prime,\onetau)_{\tau=1}^t)-\log{\Prob_{\hat{f}}}((x_\tau, x_\tau^\prime,\onetau)_{\tau=1}^t\right)\nonumber\\
 \leq&C(\epsilon, \delta, t)+2C_L\epsilon t+2\beta_1(\epsilon, \delta, t)\label{inq:squ5}\\
 =&\beta_2(\epsilon,\delta, t)+2\beta_1(\epsilon, \delta, t),
\end{align}
where $C(\epsilon, \delta, t)=8H_\sigma\bar{\sigma^\prime}^2\epsilon^2t+\sqrt{8{tB_p^2\log\frac{\pi^2t^2\mathcal{N}(\mathcal{B}_f,\epsilon, \|\cdot\|_\infty)}{6\delta}}{}}$ and $\beta_2(\epsilon, \delta, t)=C(\epsilon, \delta, t)+2C_L\epsilon t$. The inequality~\meqref{inq:squ1} follows by the fact that $(a+b)^2\leq2a^2+2b^2,\forall a,b\in\mathbb{R}$. The inequality~\meqref{inq:squ2} follows because $\|\hat{f}_{t+1}-\hat{f}\|_\infty\leq\epsilon$. The inequality~\meqref{inq:squ3} follows by Lem.~\ref{lem:p_square_bound} (with reset of `$\delta$'). The inequality~\meqref{inq:squ4} follows by that $$\log{\Prob_{\hat{f}_{t}^\mathrm{MLE}}}((x_\tau, x_\tau^\prime,\onetau)_{\tau=1}^t)\geq\log{\Prob_{{f}}}((x_\tau, x_\tau^\prime,\onetau)_{\tau=1}^t).$$ The inequality~\meqref{inq:squ5} follows by the fact that $\hat{f}_{t+1}\in\mathcal{B}^{t+1}_f$ and Lem.~\ref{lem:close_f}. 

Furthermore,
\begin{align}
   \sum_{\tau=1}^t\left(\hat{p}^{t+1}_\tau-p_\tau\right)^2&=\sum_{\tau=1}^t\left(\sigma\left(\hat{f}_{t+1}(x_\tau)-\hat{f}_{t+1}(x_\tau^\prime)\right)-\sigma\left({f}(x_\tau)-{f}(x_\tau^\prime)\right)\right)^2\\
   &\geq\sum_{\tau=1}^t{\underline{\sigma}^\prime}^2\left(\left(\hat{f}_{t+1}(x_\tau)-\hat{f}_{t+1}(x_\tau^\prime)\right)-\left({f}(x_\tau)-{f}(x_\tau^\prime)\right)\right)^2, 
\end{align}
where the inequality follows by mean value theorem. The conclusion then follows.

\section{Proof of Thm.~\ref{thm:bound_duelwise_error}}
 Before we proceed to prove Thm.~\ref{thm:bound_duelwise_error}, we first conduct a black-box analysis in Sec.~\ref{app_sec:black_box_analy} to bound the pointwise error for a generic RKHS with a generic learning scheme, which we think can be of independent interest. 
\subsection{Black-box analysis on the pointwise inference error in a generic RKHS}
\label{app_sec:black_box_analy}
Suppose we have a generic RKHS $\tilde{\mathcal{H}}$ with a generic positive semidefinite kernel function $\tilde{k}(\cdot, \cdot)$. After obtaining some information~(preference information in this paper) on a sequence $\xx_1, \xx_2,\cdot, \xx_{t-1}$, a learning scheme outputs a learnt uncertainty set, 
\begin{equation}
{\mathcal{S}}_t=\{\tilde{h}\in\mathcal{B}|\sum_{\tau=1}^{t-1}\left(\tilde{h}(\xx_\tau)-{h}(\xx_\tau)\right)^2\leq\tilde{\beta}_t\},
\end{equation}
where $\mathcal{B}$ is a function space ball with radius $\tilde{B}$ in $\tilde{\mathcal{H}}$, $h\in\mathcal{B}$ is the ground truth function and $\tilde{\beta}_t$ quantifies the size of this confidence set. Let $\tilde{\mathcal{X}}$ denote the function input set, which is assumed to be compact. We introduce the function, 
\begin{align}
\tilde{\sigma}_{t}^2(\xx) &=\tilde{k}\left(\xx, {\xx}\right)-\tilde{k}(\xx_{1:t-1}, \xx)^\top\left(\tilde{K}_{t-1}+ \lambda I\right)^{-1} \tilde{k}\left(\xx_{1:t-1}, {\xx}\right), \label{eqn:sigma_def}
\end{align}
where $\lambda$ is a positive constant and $\tilde{K}_{t-1}=(\tilde{k}(\tilde{x}_i, \tilde{x}_j))_{i\in[t-1], j\in[t-1]}$. We have the following theorem.
\begin{theorem}
    \label{thm:bound_width} 
    $\forall \hh\in\mathcal{S}_{t+1}, \xx\in\tilde{\mathcal{X}}$, we have,
    \begin{equation}
      \abs{\hh(\xx)-h(\xx)}\leq 2(\tilde{B}+\lambda^{-\nicefrac{1}{2}}\tilde{\beta}_{t+1}^{\nicefrac{1}{2}})\tilde{\sigma}_{t+1}(\xx).
    \end{equation}
\end{theorem}
\begin{proof}
    
For simplicity, we use $\phi(\xx)$ to denote the function $\kk(\xx,\cdot)$, where $\phi: \mathbb{R}^{\dd} \to \tilde{\mathcal{H}}$ maps a finite dimensional point $\xx\in\mathbb{R}^{\dd}$ to the RKHS $\tilde{\mathcal{H}}$. For simplicity, we use $h_1^\top h_2$ to denote the inner product of two functions $h_1, h_2$ from the RKHS $\tilde{\Hil}$. Therefore, $h(\xx)=\langle h, \kk(\xx, \cdot)\rangle_{\kk}=h^\top\phi(\xx)$ and
$\kk(\xx, \bar{\xx})=\langle \kk(\xx, \cdot), \kk(\bar{\xx}, \cdot)\rangle=\phi(\xx)^\top\phi(\bar{\xx})$, $\forall \xx, \bar{\xx}\in \tilde{\mathcal{X}}$. We can introduce the feature map $$\Phi_t \defeq\left[\phi(\xx_1)^\top,\hdots,\phi(\xx_{t})^\top\right]^\top,$$ 
we then get the kernel matrix $\tilde{K}_t = \Phi_t\Phi_t^\top=(\kk(\xx_i, \xx_j))_{i,j\in[t]}$, $\kk_t(\xx)= \Phi_t\phi(\xx)=(\kk(\xx, \xx_i))_{i\in[t]}$ for all $\xx\in\tilde{\mathcal{X}}$ and $h_{1:t}=\Phi_th$.

Note that when the Hilbert space $\tilde{\Hil}$ is finite-dimensional, $\Phi_t$ is interpreted as the normal finite-dimensional matrix. In the more general setting where $\tilde{\Hil}$ can be an infinite-dimensional space, $\Phi_t$ is the evaluation operator $\tilde{\Hil}\to\mathbb{R}^t$ defined as $\Phi_t h\defeq[h(\xx_1),\cdots,h(\xx_t)]^\top,\forall h\in\tilde{\Hil}$, with $\Phi_t^\top:\mathbb{R}^t\to\tilde{H}$ as its adjoint operator. For the simplicity of notation, we abuse the notation $I$ to denote the identity mapping in both the RKHS $\tilde{\Hil}$ and $\mathbb{R}^t$. The specific meaning of $I$ depends on the context. 

Since the matrices/operators $(\Phi_t^\top\Phi_t + \lambda I)$ and $(\Phi_t\Phi_t^\top + \lambda I)$ are strictly positive definite and $$(\Phi_t^\top\Phi_t + \lambda I)\Phi_t^\top = \Phi_t^\top(\Phi_t\Phi_t^\top + \lambda I),$$ we have
\begin{equation}
    \Phi_t^\top(\Phi_t\Phi_t^\top + \lambda I)^{-1} = (\Phi_t^\top\Phi_t + \lambda I)^{-1}\Phi_t^\top.
\label{eqn:dim-change}
\end{equation}
Also from the definitions above $(\Phi_t^\top\Phi_t+\lambda I)\phi(\xx)=\Phi_t^\top \kk_t(\xx) + \lambda \phi(\xx)$, and thus $\phi(\xx)=(\Phi_t^\top\Phi_t+\lambda I)^{-1}\Phi_t^\top \kk_t(\xx) + \lambda(\Phi_t^\top\Phi_t+\lambda I)^{-1}\phi(\xx)$. Hence, from Eq.~\meqref{eqn:dim-change} we deduce that
\begin{equation}
    \phi(\xx)=\Phi_t^\top(\Phi_t\Phi_t^\top+\lambda I)^{-1}\kk_t(\xx)+\lambda (\Phi_t^\top\Phi_t + \lambda I)^{-1}\phi(\xx), \label{eq:phx}
\end{equation}
which gives
\begin{equation}
    \phi(\xx)^\top\phi(\xx)= \kk_t(\xx)^\top(\Phi_t\Phi_t^\top+\lambda I)^{-1}\kk_t(\xx)+\lambda \phi(\xx)^\top(\Phi_t^\top\Phi_t + \lambda I)^{-1}\phi(\xx),
\end{equation}
by multiplying both sides of Eq.~\meqref{eq:phx} with $\phi(\xx)^\top$. 
This implies
\begin{equation}
    \lambda \phi(\xx)^\top(\Phi_t^\top\Phi_t + \lambda I)^{-1}\phi(\xx) = \kk(\xx,\xx) -\kk_t(\xx)^\top(\tilde{K}_t+\lambda I)^{-1}\kk_t(\xx) = \tilde{\sigma}_{t+1}^2(\xx),
\label{eqn:variance}
\end{equation}
where the second equality follows by the definition of $\tilde{\sigma}_{t+1}(\xx)$. 
Now observe that $\forall\hh\in\mathcal{B}$, 
\begin{align} 
&\abs{\hh(\xx)-\kk_t(\xx)^\top(\tilde{K}_t+\lambda I)^{-1}\hh_{1:t}} \\
=& \abs{\phi(\xx)^\top \hh- \phi(\xx)^\top\Phi_t^\top(\Phi_t\Phi_t^\top+\lambda I)^{-1}\Phi_t \hh}\\
=& \abs{\phi(\xx)^\top \hh-\phi(\xx)^\top(\Phi_t^\top\Phi_t + \lambda I)^{-1}\Phi_t^\top\Phi_t \hh}\label{eq:change_dim_Phi}\\
=& \abs{\phi(\xx)^\top(\Phi_t^\top\Phi_t + \lambda I)^{-1}(\Phi_t^\top\Phi_t+\lambda I)\hh-\phi(\xx)^\top(\Phi_t^\top\Phi_t + \lambda I)^{-1}\Phi_t^\top\Phi_t \hh}\label{eq:rewrite}\\
=& \abs{\lambda \phi(\xx)^\top(\Phi_t^\top\Phi_t+\lambda I)^{-1}\hh}\\
\le& \norm{\lambda(\Phi_t^T\Phi_t+\lambda I)^{-1}\phi(\xx)}_{\kk}\norm{\hh}_{\kk}\label{inq:CS}\\
=& \norm{\hh}_{\kk} \sqrt{\lambda \phi(\xx)^\top(\Phi_t^\top\Phi_t+\lambda I)^{-1}\lambda I(\Phi_t^\top\Phi_t+ \lambda I)^{-1}\phi(\xx)}\\
\le & \tilde{B} \sqrt{\lambda \phi(\xx)^\top(\Phi_t^\top\Phi_t+\lambda I)^{-1}(\Phi_t^\top\Phi_t+\lambda I)(\Phi_t^\top\Phi_t+\lambda I)^{-1}\phi(\xx)}\label{inq:add_term}\\
=&\tilde{B}\; \tilde{\sigma}_{t+1}(\xx), \label{eq:bound_res_f} 
\end{align}
where the equality~\meqref{eq:change_dim_Phi} uses Eq.~\meqref{eqn:dim-change}, the inequality~\meqref{inq:CS} is by Cauchy-Schwartz, the inequality~\meqref{inq:add_term} follows by the assumption that $\|\tilde{h}\|_{\tilde{k}}\leq\tilde{B}$ and that $\Phi_t^\top\Phi_t$ is positive semidefinite, and the equality~\meqref{eq:bound_res_f} is from Eq.~\meqref{eqn:variance}. 
We define $\Delta_{1:t}=\tilde{h}_{1:t}-{h}_{1:t}$,
\begin{align}
   & |{\kk_t(\xx)^\top(\tilde{K}_t+\lambda I)^{-1}\Delta_{1:t}}|\\ =&\abs{\phi(\xx)^\top\Phi_t^\top(\Phi_t\Phi_t^\top + \lambda I)^{-1}\Delta_{1:t}}\\
=& \abs{\phi(\xx)^\top(\Phi_t^\top\Phi_t + \lambda I)^{-1}\Phi_t^\top\Delta_{1:t}} \label{eq:change_dim_aga}\\
\le &\norm{(\Phi_t^\top\Phi_t + \lambda I)^{-1/2}\phi(\xx)}_{\kk} \norm{(\Phi_t^\top\Phi_t + \lambda I)^{-1/2}\Phi_t^\top\Delta_{1:t}}_{\kk}\label{inq:CS_del}\\
= &\sqrt{\phi(\xx)^\top(\Phi_t^\top\Phi_t + \lambda I)^{-1}\phi(\xx)}\sqrt{(\Phi_t^\top\Delta_{1:t})^\top(\Phi_t^\top\Phi_t + \lambda I)^{-1}\Phi_t^\top\Delta_{1:t}}\\
=&\lambda^{-1/2}\tilde{\sigma}_{t+1}(\xx)\sqrt{\Delta_{1:t}^\top\Phi_t\Phi_t^\top(\Phi_t\Phi_t^\top +\lambda I)^{-1}\Delta_{1:t}}\label{eq:change_dim_sigma}\\
=&\lambda^{-1/2}\tilde{\sigma}_{t+1}(\xx)\sqrt{\Delta_{1:t}^\top \tilde{K}_t(\tilde{K}_t+\lambda I)^{-1}\Delta_{1:t}}\\
\leq&\lambda^{-1/2}\tilde{\sigma}_{t+1}(\xx)\sqrt{\Delta_{1:t}^\top\Delta_{1:t}}\\
\leq&\lambda^{-1/2}\tilde{\beta}_{t+1}^{1/2}\tilde{\sigma}_{t+1}(\tilde{x}) \label{eq:bound_int_diff}
\end{align}
where the equality~\meqref{eq:change_dim_aga} is from Eq.~\meqref{eqn:dim-change}, the inequality~\meqref{inq:CS_del} is by Cauchy-Schwartz and the equality~\meqref{eq:change_dim_sigma} uses both Eq.~\meqref{eqn:dim-change} and Eq.~\meqref{eqn:variance}. We can finally derive,
\begin{align}
    &\left|\hh(\xx)-h(\xx)\right| \\
    =&\left|{\kk_t(\xx)^\top(\tilde{K}_t+\lambda I)^{-1}(\tilde{h}_{1:t}-h_{1:t})} -\left({h(\xx)-\kk_t(\xx)^T(\tilde{K}_t+\lambda I)^{-1}h_{1:t}}\right)+\left({\hh(\xx)-\kk_t(\xx)^\top(\tilde{K}_t+\lambda I)^{-1}\hh_{1:t}}\right)\right|\label{eq:split}\\
    \le & \left|{\kk_t(\xx)^\top(\tilde{K}_t+\lambda I)^{-1}(\tilde{h}_{1:t}-h_{1:t})}\right| + \left|{h(\xx)-\kk_t(\xx)^T(\tilde{K}_t+\lambda I)^{-1}h_{1:t}}\right| + \left|{\hh(\xx)-\kk_t(\xx)^\top(\tilde{K}_t+\lambda I)^{-1}\hh_{1:t}}\right|\label{inq:triangle}\\
    \le& \Big(2\tilde{B} + \lambda^{-1/2}\tilde{\beta}_{t+1}^{1/2}\Big)\tilde{\sigma}_{t+1}(\xx),
\end{align}
where the equality~\meqref{eq:split} follows by splitting, the inequality~\meqref{inq:triangle} follows by triangle inequality, the last inequality follows by combining the inequality~\meqref{eq:bound_res_f} and the inequality~\meqref{eq:bound_int_diff}. The conclusion then follows.
\end{proof}
\begin{remark}
   The proof idea is inspired by the proof of Thm. 2 in~\cite{chowdhury2017kernelized_arxiv}.  
\end{remark}

\subsection{Main proof of Thm.~\ref{thm:bound_duelwise_error}}
We set the generic RKHS $\tilde{\Hil}$ to be the augmented RKHS with the additive kernel function $k^{ff^\prime}$, the function space ball to be $\mathcal{B}_{ff^\prime}$, $\tilde{B}=2B$ and the confidence set as,
\begin{equation}
{\mathcal{S}}_t\defeq\left\{\ff(x)-\ff(x^\prime)|\ff\in\mathcal{B}_f\nonumber, \sum_{\tau=1}^{t-1}\big((\ff(x_\tau)-\ff(x_\tau^\prime))-\left({f}(x_\tau)-{f}(x^\prime_\tau)\right)\big)^2
\leq\beta(\epsilon, \nicefrac{\delta}{2}, t-1)\right\}\subset\mathcal{B}_{ff^\prime}.
\label{eq:lifted_conf_set} 
\end{equation}
The desired result then follows by applying Thm.~\ref{thm:bound_width}.    

\section{Proof of Lem.~\ref{thm:inner_finite_red}}
It suffices to prove that for any feasible solution of Prob.~\meqref{eqn:generic_inner_prob}, we can find a corresponding feasible solution of Prob.~\meqref{eqn:reform_inner_prob_to_fin} with the same objective value and that the inverse direction also holds.
\begin{enumerate}
    \item In this part, we first show that for any feasible solution of Prob.~\meqref{eqn:generic_inner_prob}, we can find a corresponding feasible solution of Prob.~\meqref{eqn:reform_inner_prob_to_fin} with the same objective value. Let $\tilde{f}$ be a feasible solution of Prob.~\meqref{eqn:generic_inner_prob}. We construct $\tilde{Z}_{0:t}=(\tilde{f}(x_\tau))_{\tau=0}^t$ and $\tilde{z}=\tilde{f}(x)$. Consider the minimum-norm interpolation problem, 
    \begin{equation}
    \label{eqn:inter_prob}
        \begin{aligned}
           \min_{s\in\mathcal{B}_f}&\;\;\|s\|^2\\
           \textrm{subject to}&\;\; s(x_\tau)=\tilde{z}_\tau, \forall\tau\in\{0\}\cup[t],\\
           &\;\;s(x)=\tilde{z}.
        \end{aligned}
    \end{equation}
   By representer theorem, the Prob.~\meqref{eqn:inter_prob} admits an optimal solution with the form $\alpha^\top k_{0:t,x}(\cdot)$, where $k_{0:t,x}\defeq (k(w,\cdot))_{w\in\{x_0,\cdots,x_t,x\}}$. So Prob.~\meqref{eqn:inter_prob} can be reduced to  
     \begin{equation}
    \label{eqn:inter_prob_fin}
        \begin{aligned}
           \min_{\alpha\in\mathbb{R}^{t+2}}&\;\;\alpha^\top K_{0:t,x}\alpha\\
           \textrm{subject to}&\;\;K_{0:t,x}\alpha=\left[\begin{array}{l}
\tilde{Z}_{0:t} \\
\tilde{z}
\end{array}\right]. 
        \end{aligned}
    \end{equation}
Hence, by solving Prob.~\meqref{eqn:inter_prob_fin}, we can derive the minimum norm square with interpolation constraints as 
$$
\left[\begin{array}{l}
\tilde{Z}_{0:t} \\
\tilde{z}
\end{array}\right]^{\top}K_{0:t,x}^{-1}\left[\begin{array}{l}
\tilde{Z}_{0:t} \\
\tilde{z}
\end{array}\right]. 
$$
Since $\tilde{f}$ itself is an interpolant by construction of $(\tilde{Z}_{0:t},\tilde{z})$. We have
$$
\left[\begin{array}{l}
\tilde{Z}_{0:t} \\
\tilde{z}
\end{array}\right]^{\top}K_{0:t,x}^{-1}\left[\begin{array}{l}
\tilde{Z}_{0:t} \\
\tilde{z}
\end{array}\right]\leq \|\tilde{f}\|^2\leq B^2. 
$$
And since the log-likelihood only depends on $\tilde{Z}_{0:t}$, it holds that $$
\ell(\tilde{Z}_{0:t}|\mathcal{D}_t)=\ell_t(\tilde{f})\geq \ell_t(\hat{f}^\mathrm{MLE}_t)-\beta_1(\epsilon, \delta, t).
$$ 
And the objectives satisfy,
$$
\tilde{z}-\tilde{z}_t=\tilde{f}(x)-\tilde{f}(x_t).
$$
Therefore, $(\tilde{Z}_{0:t},\tilde{z})$ is a feasible solution for Prob.~\meqref{eqn:reform_inner_prob_to_fin} with the same objective as $\tilde{f}$ for Prob.~\meqref{eqn:generic_inner_prob}.

\item We then show that for any feasible solution of Prob.~\meqref{eqn:reform_inner_prob_to_fin}, we can find a corresponding feasible solution of Prob.~\meqref{eqn:generic_inner_prob} with the same objective value. Let $(Z_{0:t}, z)$ be a feasible solution of Prob.~\meqref{eqn:reform_inner_prob_to_fin}. We construct 
$$
\tilde{f}_z = \left[\begin{array}{l}
{Z}_{0:t} \\
{z}
\end{array}\right]^{\top}K_{0:t,x}^{-1}k_{0:t,x}(\cdot).
$$
Hence, 
$$
\|\tilde{f}_z\|^2 =\left[\begin{array}{l}
{Z}_{0:t} \\
{z}
\end{array}\right]^{\top}K_{0:t,x}^{-1}\left[\begin{array}{l}
{Z}_{0:t} \\
{z}
\end{array}\right] \leq B^2.
$$
And it can be checked that 
$\tilde{f}_z(x_\tau)=z_\tau,\forall \tau\in\{0\}\cup[t]$ and $\tilde{f}_z(x)=z$.
So $\ell_t(\tilde{f}_z)=\ell(Z_{0:t}|\mathcal{D}_t)\geq\ell_t(\hat{f}^\mathrm{MLE}_t)-\beta_1(\epsilon, \delta, t)$. And the objectives satisfy $\tilde{f}_z(x)-\tilde{f}_z(x_t)=z-z_t$.
So it is proved that for any feasible solution of Prob.~\meqref{eqn:reform_inner_prob_to_fin}, we can find a corresponding feasible solution of Prob.~\meqref{eqn:generic_inner_prob} with the same objective value.
\end{enumerate}
The desired result then follows.

\rev{
\section{Elaboration on Remark~\ref{remark_choice_B}}\label{app_sec:ela_on_hyp} By assumption $2.2$, we assume that there exists a large enough constant $B$ that upper bounds the norm of the ground-truth black-box function $f$. However, the exact value of this upper bound may be unknown to us in practice, while the execution of our algorithm relies on the knowledge of $B$ (in Problem (23), $B$ is a key parameter). So we need to guess the value of $B$. Suppose our guess is $\hat{B}$. It is possible that $\hat{B}$ is even smaller than the ground-truth function norm $\|f\|$. To detect this wrong guess, we observe that, with the correct setting of $B$ such that $B\geq\|f\|$, we have that by Thm. 3.1 and the definition of maximum likelihood estimate, with high probability,
$$
\ell_t(\hat{f}^{\mathrm{MLE}}_{t|B})\geq\ell_t(f)\geq\ell_t(\hat{f}^{\mathrm{MLE}}_{t|B})-\beta_1(\epsilon, \delta, t|B),
$$
where $\hat{f}^{\mathrm{MLE}}_{t|B}$ is the maximum likelihood estimate function with function norm bound $B$ and  $\beta_1(\epsilon, \delta, t|B)$ is the corresponding parameter as defined in Thm. 3.1 with norm bound $B$. We also have $2B$ is a valid upper bound on $\|f\|$ and thus,     
$$
\ell_t(\hat{f}^{\mathrm{MLE}}_{t|2B})\geq\ell_t(f)\geq\ell_t(\hat{f}^{\mathrm{MLE}}_{t|2B})-\beta_1(\epsilon, \delta, t|2B).
$$
Hence, 
$$
\ell_t(\hat{f}^{\mathrm{MLE}}_{t|B})\geq\ell_t(f)\geq\ell_t(\hat{f}^{\mathrm{MLE}}_{t|2B})-\beta_1(\epsilon, \delta, t|2B).
$$
That is to say, $\ell_t(\hat{f}^{\mathrm{MLE}}_{t|B})$ needs to be greater than or equal to $\ell_t(\hat{f}^{\mathrm{MLE}}_{t|2B})-\beta_1(\epsilon, \delta, t|2B)$ when $B$ is a valid upper bound on $\|f\|$. 

Therefore, we can use the heuristic: every time we find that
$$
\ell_t(\hat{f}^{\mathrm{MLE}}_{t|\hat{B}})<\ell_t(\hat{f}^{\mathrm{MLE}}_{t|2\hat{B}})-\beta_1(\epsilon, \delta, t|2\hat{B}),
$$
we double the upper bound guess $\hat{B}$.
}

\rev{
\section{Jointly optimize $x$, $Z_{0,t}$ and $z$ for the problem~\meqref{eqn:reform_inner_prob_to_fin}.}\label{app_sec:joint_opt} For medium-dimensional problems ($d>4$), we can jointly optimize $x$, $Z_{0:t}$, and $z$ by a nonlinear programming solver from multiple random initial conditions. That is, we can also treat $x$ in the problem (23) as an optimization variable. In this way, we lose convexity but only need to solve the problem (23) for only \emph{once} in each step $t$. 

More specifically, we solve the optimization problem~\meqref{eqn:reform_inner_prob_to_fin_joint}.

\begin{equation}    
\begin{aligned}
    \label{eqn:reform_inner_prob_to_fin_joint}
\max_{x\in\mathbb{R}^d, Z_{0:t}\in\mathbb{R}^{t+1}, z\in\mathbb{R}}&\quad z- z_t \\
   \text{subject to}&\quad \left[\begin{array}{l}
Z_{0:t} \\
z
\end{array}\right]^{\top}K_{0:t,x}^{-1}\left[\begin{array}{l}
Z_{0:t} \\
z
\end{array}\right]   \leq B^2, \\
   &\quad \ell(Z_{0:t}|\mathcal{D}_t)\geq \ell_t(\hat{f}^\mathrm{MLE}_t)-\beta_1(\epsilon, \delta, t),
\end{aligned}
\end{equation}

The only constraint that involves $x$ is 
\begin{equation}
\quad \left[\begin{array}{l}
Z_{0:t} \\
z
\end{array}\right]^{\top}K_{0:t,x}^{-1}\left[\begin{array}{l}
Z_{0:t} \\
z
\end{array}\right]   \leq B^2. 
\end{equation}
Applying matrix inversion, we derive that the left-hand side is equal to,
\begin{equation}
Z_{0:t}^\top K_{0:t}^{-1}Z_{0:t}+\frac{1}{k(x,x)-k_t(x)^\top K_{0:t}^{-1}k_t(x)} \quad \left[\begin{array}{l}
Z_{0:t} \\
z
\end{array}\right]^{\top}  \left[\begin{array}{l}
K_{0:t}^{-1}k_t(x) \\
-1
\end{array}\right] \left[\begin{array}{l}
K_{0:t}^{-1}k_t(x) \\
-1
\end{array}\right]^\top\left[\begin{array}{l}
Z_{0:t} \\
z
\end{array}\right],    
\end{equation}
where $k_t(x)\defeq(k(x_\tau, x))_{\tau=0}^t$.

We can then apply a nonlinear programming solver such as Ipopt to solve the problem~\meqref{eqn:reform_inner_prob_to_fin_joint} from randomly sampled initial points. Then the best converged solution is set to be the next sample point $x_t$.  
}

\rev{
\section{Extension to the multiple-choice setting}
\label{app_sec:ext_multi_choice}
In this paper, we mainly consider the setting where human expresses preference over only two choices, because of its low cognitive burden to the human user and simplicity of theoretical analysis.
However, we can extend POP-BO to the multiple-choice setting where human can compare multiple choices and express the favorite one. 

Suppose that in each step $\tau$, we aim to generate a batch of $q$ points. Then we can mix the new batch with the old batch generated in step $\tau-1$, and query the comparison oracle to report the favorite point among the $2q$ points. 

Firstly, the confidence set of functions can be similarly constructed using the likelihood ratio idea and the multiple-choice probabilistic preference model as in (Astudilo et al. 2023),
\begin{equation}
\Prob\leb x_r \text{ is the favorite} \rib = \frac{e^{f(x_r)}}{\sum_{x\in\{\text{last batch and the new batch}\}}e^{f(x)}}.
\end{equation}

Secondly, to generate the new batch, the basic idea is that we can apply a `bootstrap'-type technique. More specifically, we can sequentially generate the new batch $x^{1}, x^{2}, \cdots, x^{q}$. When generating the new point $x^{r+1}$, we maximize its corresponding optimistic advantage of $z^{r+1}$ as compared to the maximum of $z_{t-q+1:t}, z^1, \cdots, z^r$ by solving a similar problem to Problem (23). That is, we solve the Problem~\meqref{eqn:reform_inner_prob_to_fin_batch} to generate the new point $x^{r+1}$ in the same batch, 
\begin{equation}    
\begin{aligned}
    \label{eqn:reform_inner_prob_to_fin_batch}
\max_{x\in\mathbb{R}^d,z\in\mathbb{R}, z^{1:r}\in\mathbb{R}^r, Z_{0:t}\in\mathbb{R}^{t+1} }&\quad z- \max\{z_{t-q+1}, \cdots, z_t, z^1, \cdots, z^r\}\\
   \text{subject to}&\quad \left[\begin{array}{l}
Z_{0:t} \\
z^{1:r}\\
z
\end{array}\right]^{\top}K_{0:t,x^{1:r},x}^{-1}\left[\begin{array}{l}
Z_{0:t} \\
z^{1:r} \\
z
\end{array}\right]   \leq B^2, \\
   &\quad \ell(Z_{0:t}|\mathcal{D}_t)\geq \ell_t(\hat{f}^\mathrm{MLE}_t)-\beta_t,
\end{aligned}
\end{equation}
which is equivalent to 
\begin{equation}    
\begin{aligned}
    \label{eqn:reform_inner_prob_to_fin_batch_refor}
\max_{x\in\mathbb{R}^d,z\in\mathbb{R}, v\in\mathbb{R}, z^{1:r}\in\mathbb{R}^r, Z_{0:t}\in\mathbb{R}^{t+1} }&\quad z- v\\
   \text{subject to}&\quad \left[\begin{array}{l}
Z_{0:t} \\
z^{1:r}\\
z
\end{array}\right]^{\top}K_{0:t,x^{1:r},x}^{-1}\left[\begin{array}{l}
Z_{0:t} \\
z^{1:r} \\
z
\end{array}\right]   \leq B^2, \\
   &\quad \ell(Z_{0:t}|\mathcal{D}_t)\geq \ell_t(\hat{f}^\mathrm{MLE}_t)-\beta_t,\\
   &\quad v\geq z_{t-i+1}, i\in[q],\\
   &\quad v\geq z^j, j\in[r],
\end{aligned}
\end{equation}
by introducing an auxiliary variable $v\in\mathbb{R}$. Problem~\meqref{eqn:reform_inner_prob_to_fin_batch_refor} can be efficiently solved by the nonlinear programming solver Ipopt. 

}


\section{Proof of Thm.~\ref{thm:regret_bound_general_kernel}}
To prepare for the following analysis, we first give a useful lemma.
   \begin{lemma}[Lemma 4,~\cite{chowdhury2017kernelized_arxiv}]
\begin{equation}
    \label{lem:bound_cumu_sd}
\sum_{t=1}^T {\sigma}^{ff^\prime}_{t}\left((x_t, x_t^\prime)\right) \leq \sqrt{4(T+2) {\gamma}^{ff^\prime}_{T}},
\end{equation}
\end{lemma} 
where $\sigma^{ff^\prime}_t$ is as defined in Eq.~\meqref{eqn:ff_sigma_def} and ${\gamma}^{ff^\prime}_{T}$ is as defined in Eq.~\meqref{eqn:info_gain_def}. 
\begin{proof}
   Apply the Lemma 4 in~\cite{chowdhury2017kernelized_arxiv} by setting the kernel function as $k^{ff^\prime}$.   
\end{proof}

For convenience, we use $\beta_t$ to denote $\beta(\epsilon, \nicefrac{\delta}{2}, t)$.
We can then analyze the regret of the optimistic algorithm. 
\begin{align*}
 R_T=& \sum_{t=1}^T[\tf(x^\star)-\tf(x_t)] \\
 =&\sum_{t=1}^T[(\tf(x^\star)-\tf(x_t^\prime))-(\tf(x_t)-\tf(x_t^\prime))]\\
 \leq &\sum_{t=1}^T[(\ff_{t}(x_t)-\ff_t(x_t^\prime))-(\tf(x_t)-\tf(x_t^\prime))]\\
\leq & \sum_{t=1}^T 2(2B+\lambda^{-\nicefrac{1}{2}}\beta_t^{\nicefrac{1}{2}})\sigma^{ff^\prime}_{t}((x_t, x_t^\prime)), 
\end{align*}
where the first inequality follows by the optimality of $(x_t, \ff_t)$ for the optimization problem in line~\ref{alg_line:opt_get_x} of the Alg.~\ref{alg:opt_pref_BO}, and the second inequality follows by Thm.~\ref{thm:bound_duelwise_error}~(Note that $\beta(\epsilon,\nicefrac{\delta}{2}, t-1)\leq\beta_t=\beta(\epsilon,\nicefrac{\delta}{2},t)$). Hence, 
\begin{align*}
 R_T\leq &\sum_{t=1}^T 2(2B+\lambda^{-\nicefrac{1}{2}}\beta_t^{\nicefrac{1}{2}})\sigma^{ff^\prime}_{t}((x_t, x_t^\prime))\\
 \leq& 2(2B+\lambda^{-\nicefrac{1}{2}}\beta_T^{\nicefrac{1}{2}})\sum_{t=1}^T \sigma^{ff^\prime}_{t}((x_t, x_t^\prime))\\
 \leq&   2(2B+\lambda^{-\nicefrac{1}{2}}\beta_T^{\nicefrac{1}{2}})\sqrt{4(T+2) \gamma_{T}^{ff^\prime}}\\
= &  \mathcal{O}\left(\sqrt{\beta_TT\gamma^{ff^\prime}_{T}}\right).
\end{align*}

\section{Proof of Thm.~\ref{thm:solution_report_and_conv_rate}}
We have
\begin{align*}
\tf(x^\star)-\tf(x_{t^\star}) =&(\tf(x^\star)-\tf(x_{t^\star}^\prime))-(\tf(x_{t^\star})-\tf(x_{t^\star}^\prime))\\
 \leq &(\ff_{t^\star}(x_{t^\star})-\ff_{t^\star}(x_{t^\star}^\prime))-(\tf(x_{t^\star})-\tf(x_{t^\star}^\prime))\\
\leq & 2(2B+\lambda^{-\nicefrac{1}{2}}\beta_{t^\star}^{\nicefrac{1}{2}})\sigma^{ff^\prime}_{t^\star}((x_{t^\star}, x_{t^\star}^\prime)), 
\end{align*}
where $\sigma^{ff^\prime}_{t^\star}$ is as given in Eq.~\meqref{eqn:ff_sigma_def} with the kernel function as $k^{ff^\prime}((x_1, x_1^\prime), (x_2, x_2^\prime))= k(x_1, x_2)+k(x_1^\prime, x_2^\prime)$ and $\beta_{t^\star}=\beta(\epsilon, \nicefrac{\delta}{2}, t^\star)$. Furthermore, by the definition of $t^\star$,  
\begin{align*}
 2(2B+\lambda^{-\nicefrac{1}{2}}\beta_{t^\star}^{\nicefrac{1}{2}})\sigma^{ff^\prime}_{t^\star}((x_{t^\star}, x_{t^\star}^\prime))\leq &\frac{1}{T}\sum_{t=1}^T 2(2B+\lambda^{-\nicefrac{1}{2}}\beta_t^{\nicefrac{1}{2}})\sigma^{ff^\prime}_{t}((x_t, x_t^\prime))\\ 
 \leq& \frac{2}{T}(2B+\lambda^{-\nicefrac{1}{2}}\beta_T^{\nicefrac{1}{2}})\sum_{t=1}^T \sigma^{ff^\prime}_{t}((x_t, x_t^\prime))\\
 \leq&   \frac{2}{T}(2B+\lambda^{-\nicefrac{1}{2}}\beta_T^{\nicefrac{1}{2}})\sqrt{4(T+2) \gamma^{ff^\prime}_{T}}\\
= &  \mathcal{O}\left(\frac{\sqrt{\beta_T\gamma^{ff^\prime}_{T}}}{\sqrt{T}}\right).
\end{align*}
The conclusion then follows.

\section{Commonly used specific kernel functions}
\label{app:spec_kerfuncs}
\begin{itemize}
\item{Linear}: $$k(x,\bar{x})=x^\top\bar{x}.$$
\item{Squared Exponential (SE)}:  $$k(x,\bar{x}) =\sigma_\mathrm{SE}^2\exp{\left\{-\frac{\norm{x-\bar{x}}^2}{l^2}\right\}},$$
where $\sigma_\mathrm{SE}^2$ is the variance parameter and $l$ is the lengthscale parameter.
\item{Mat\'ern}: 
$$
k(x,\bar{x})=\frac{2^{1-\nu}}{\Gamma(\nu)}\left(\sqrt{2 \nu} \frac{\norm{x-\bar{x}}}{\rho}\right)^{\nu} K_{\nu}\left(\sqrt{2 \nu} \frac{\norm{x-\bar{x}}}{\rho}\right),    
$$
where $\rho$ and $\nu$ are the two positive parameters of the kernel function, $\Gamma$ is the gamma function, and $K_{\nu}$ is the modified Bessel function of the second kind. $\nu$ captures the smoothness of the kernel function. 
\end{itemize}

\section{Proof of Thm.~\ref{thm:kern_spec_bound}}
\label{app_sec:kern_spec_Rbound}
Recall that $$\beta(\epsilon, \nicefrac{\delta}{2}, t)=\frac{\underline{\sigma^\prime}^2}{H_\sigma}\left(\beta_2(\epsilon,\delta, t)+2\beta_1(\epsilon, \delta, t)\right)=\mathcal{O}\left(\sqrt{t\log\frac{t\mathcal{N}(\mathcal{B}_f, \epsilon,\|\cdot\|_\infty)}{\delta}}+\epsilon t+\epsilon^2t\right).$$ We pick $\epsilon=\nicefrac{1}{T}$, and can thus derive,
$$
\beta_T=\beta(T^{-1}, \nicefrac{\delta}{2}, T)=\mathcal{O}\left(\sqrt{T\log\frac{T\mathcal{N}(\mathcal{B}_f, T^{-1},\|\cdot\|_\infty)}{\delta}}\right).
$$
\begin{enumerate}
    \item $k$ is a linear kernel, then the corresponding RKHS is a finite-dimensional space and $\log\mathcal{N}(\mathcal{B}_f, T^{-1},\|\cdot\|_\infty)=\mathcal{O}\left(\log\frac{1}{\epsilon}\right)=\mathcal{O}\left(\log T\right)$~(see, e.g.,~\cite{wu2017lecture}). The corresponding $k^{ff^\prime}((x, x^\prime), (y, y^\prime))=x^\top y+{x^\prime}^\top y^\prime=\langle (x, x^\prime), (y, y^\prime)\rangle$, which is also linear. Thus, by Thm. 5 in~\cite{srinivas2012information},
    $$
    \gamma_T^{ff^\prime}=\mathcal{O}(\log T). 
    $$
    Hence, 
    $$
    R_T=\mathcal{O}\left((T\log T)^{1/4+1/2}\right)=\mathcal{O}\left(T^{\nicefrac{3}{4}}(\log T)^{\nicefrac{3}{4}}\right). 
    $$

 \item $k$ is a squared exponential kernel, then $\log\mathcal{N}(\mathcal{B}_f, T^{-1},\|\cdot\|_\infty)=\mathcal{O}\left((\log\frac{1}{\epsilon})^{d+1}\right)=\mathcal{O}\left((\log T)^{d+1}\right)$~(Example 4,~\cite{zhou2002covering}). By Thm. 4 in~\cite{kandasamy2015high}, we have,
    $$
    \gamma_T^{ff^\prime}=\mathcal{O}((\log T)^{d+1}). 
    $$
    Hence, 
    $$
    R_T=\mathcal{O}\left(T^{\nicefrac{3}{4}}(\log T)^{\nicefrac{3}{4}(d+1)}\right). 
    $$

 \item $k$ is a Mat\'ern kernel. Lem.~3 in~\cite{bull2011convergence} implies the equivalence between RKHS and Sobolev Hilbert space. We can then apply the rich results on the bound of covering number of Sobolev Hilbert space~\cite{edmunds1996function}. So $\log\mathcal{N}(\mathcal{B}_f, T^{-1},\|\cdot\|_\infty)=\mathcal{O}\left((\frac{1}{\epsilon})^{\nicefrac{d}{\nu}}\log\frac{1}{\epsilon}\right)=\mathcal{O}\left(T^{\nicefrac{d}{\nu}}\log T\right)$~(by combing the lower bound in Thm. 5.1~\cite{xu2022lower} and the convergence rate in Thm.~1~\cite{bull2011convergence}). By Thm. 4 in~\cite{kandasamy2015high}, we have,
    $$
    \gamma_T^{ff^\prime}=\mathcal{O}\left(T^{\frac{d(d+1)}{2\nu+d(d+1)}}\log T\right). 
    $$
    Hence, 
    $$
    R_T=\mathcal{O}\left(T^{\nicefrac{3}{4}}(\log T)^{\nicefrac{3}{4}}T^{\frac{d}{\nu}\left(\frac{1}{4}+\frac{d+1}{4+2(d+1)\nicefrac{d}{\nu}}\right)}\right)\leq\mathcal{O}\left(T^{\nicefrac{3}{4}}(\log T)^{\nicefrac{3}{4}}T^{\frac{1}{4}\frac{d(d+2)}{\nu}}\right). 
    $$
\end{enumerate}

\section{Empirical Evidence for the Order of The Cumulative Regret}
\label{app_sec:Rloglog}
\begin{figure}[h!]
    \centering
\begin{tikzpicture}

\definecolor{darkgray176}{RGB}{176,176,176}
\definecolor{green}{RGB}{0,128,0}
\definecolor{lightgray204}{RGB}{204,204,204}

\begin{axis}[
width=\scplotwidth,
height=\scplotheight,
legend cell align={left},
legend columns=3,
legend style={
  fill opacity=0.8,
  draw opacity=1,
  text opacity=1,
  at={(0.5,-0.52)},
  anchor=south,
  draw=lightgray204
},
log basis x={10},
log basis y={10},
tick align=outside,
tick pos=left,
x grid style={darkgray176},
xlabel={Step},
xmin=0.822340159426889, xmax=60.8020895329329,
xmode=log,
xtick style={color=black},
xtick={0.01,0.1,1,10,100,1000},
xticklabels={
  \(\displaystyle {10^{-2}}\),
  \(\displaystyle {10^{-1}}\),
  \(\displaystyle {10^{0}}\),
  \(\displaystyle {10^{1}}\),
  \(\displaystyle {10^{2}}\),
  \(\displaystyle {10^{3}}\)
},
y grid style={darkgray176},
ylabel={Cumulative Regret},
ymin=0.822340159426889, ymax=60.8020895329329,
ymode=log,
ytick style={color=black},
ytick={0.01,0.1,1,10,100,1000},
yticklabels={
  \(\displaystyle {10^{-2}}\),
  \(\displaystyle {10^{-1}}\),
  \(\displaystyle {10^{0}}\),
  \(\displaystyle {10^{1}}\),
  \(\displaystyle {10^{2}}\),
  \(\displaystyle {10^{3}}\)
}
]
\addplot [blue, mark=triangle, mark size=1.5, mark options={solid,fill=none}]
table {%
1 1
2.00000023841858 2.00000023841858
2.99999976158142 2.99999976158142
4.00000095367432 4.00000095367432
5.00000095367432 5.00000095367432
6.00000047683716 6.00000047683716
6.99999952316284 6.99999952316284
8 8
8.99999904632568 8.99999904632568
10 10
11 11
11.9999990463257 11.9999990463257
13 13
14.0000009536743 14.0000009536743
15.0000019073486 15.0000019073486
16.0000019073486 16.0000019073486
17.0000019073486 17.0000019073486
18 18
19.0000019073486 19.0000019073486
20.0000019073486 20.0000019073486
20.9999980926514 20.9999980926514
22.0000019073486 22.0000019073486
23 23
24 24
25.0000019073486 25.0000019073486
26.0000019073486 26.0000019073486
27.0000038146973 27.0000038146973
28.0000038146973 28.0000038146973
29.0000038146973 29.0000038146973
29.9999980926514 29.9999980926514
31.0000038146973 31.0000038146973
31.9999980926514 31.9999980926514
33 33
33.9999961853027 33.9999961853027
35.0000038146973 35.0000038146973
36.0000038146973 36.0000038146973
37.0000038146973 37.0000038146973
37.9999961853027 37.9999961853027
38.9999961853027 38.9999961853027
39.9999961853027 39.9999961853027
41.0000038146973 41.0000038146973
42 42
43.0000038146973 43.0000038146973
43.9999961853027 43.9999961853027
45.0000038146973 45.0000038146973
46.0000038146973 46.0000038146973
47.0000076293945 47.0000076293945
48.0000038146973 48.0000038146973
48.9999961853027 48.9999961853027
49.9999961853027 49.9999961853027
};
\addlegendentry{$T$}
\addplot [black, mark=+, mark size=1.5, mark options={solid,fill=none}]
table {%
1 1
2.00000023841858 1.41421341896057
2.99999976158142 1.73205101490021
4.00000095367432 2.00000023841858
5.00000095367432 2.23606824874878
6.00000047683716 2.44948983192444
6.99999952316284 2.64575123786926
8 2.82842707633972
8.99999904632568 2.99999976158142
10 3.16227769851685
11 3.31662535667419
11.9999990463257 3.46410131454468
13 3.60555124282837
14.0000009536743 3.74165797233582
15.0000019073486 3.8729829788208
16.0000019073486 4.00000095367432
17.0000019073486 4.12310457229614
18 4.24264001846313
19.0000019073486 4.35889959335327
20.0000019073486 4.47213697433472
20.9999980926514 4.58257627487183
22.0000019073486 4.69041585922241
23 4.79583072662354
24 4.89898014068604
25.0000019073486 5.00000095367432
26.0000019073486 5.09901905059814
27.0000038146973 5.19615125656128
28.0000038146973 5.29150152206421
29.0000038146973 5.38516521453857
29.9999980926514 5.47722625732422
31.0000038146973 5.5677638053894
31.9999980926514 5.65685319900513
33 5.74456262588501
33.9999961853027 5.83095169067383
35.0000038146973 5.91607999801636
36.0000038146973 6.00000047683716
37.0000038146973 6.08276271820068
37.9999961853027 6.16441297531128
38.9999961853027 6.24499702453613
39.9999961853027 6.32455587387085
41.0000038146973 6.40312480926514
42 6.48074054718018
43.0000038146973 6.55743789672852
43.9999961853027 6.63324928283691
45.0000038146973 6.70820426940918
46.0000038146973 6.78233051300049
47.0000076293945 6.85565519332886
48.0000038146973 6.92820358276367
48.9999961853027 6.99999952316284
49.9999961853027 7.07106828689575
};
\addlegendentry{$\sqrt{T}$}
\addplot [green, mark=triangle, mark size=1.5, mark options={solid,rotate=180,fill=none}]
table {%
1 1
2.00000023841858 2.11954355239868
2.99999976158142 2.82055950164795
4.00000095367432 3.46154832839966
5.00000095367432 4.09937953948975
6.00000047683716 4.85304069519043
6.99999952316284 5.3165168762207
8 5.92114400863647
8.99999904632568 6.50113582611084
10 6.89976501464844
11 7.42007923126221
11.9999990463257 7.88232564926147
13 8.22537422180176
14.0000009536743 8.67196941375732
15.0000019073486 8.9763240814209
16.0000019073486 9.31973838806152
17.0000019073486 9.8336877822876
18 10.1796588897705
19.0000019073486 10.5852499008179
20.0000019073486 10.9088230133057
20.9999980926514 11.228687286377
22.0000019073486 11.4863357543945
23 11.7333755493164
24 11.9972286224365
25.0000019073486 12.1787147521973
26.0000019073486 12.3661594390869
27.0000038146973 12.5585889816284
28.0000038146973 12.8638210296631
29.0000038146973 13.2024221420288
29.9999980926514 13.4186420440674
31.0000038146973 13.617075920105
31.9999980926514 13.851282119751
33 14.0058584213257
33.9999961853027 14.2056875228882
35.0000038146973 14.4871625900269
36.0000038146973 14.6607732772827
37.0000038146973 14.7686519622803
37.9999961853027 14.925929069519
38.9999961853027 15.0491275787354
39.9999961853027 15.1816511154175
41.0000038146973 15.3011589050293
42 15.4674692153931
43.0000038146973 15.577075958252
43.9999961853027 15.7077741622925
45.0000038146973 15.8235301971436
46.0000038146973 15.9738483428955
47.0000076293945 16.0823574066162
48.0000038146973 16.2952423095703
48.9999961853027 16.403844833374
49.9999961853027 16.5425758361816
};
\addlegendentry{$R_T$ of POP-BO}

\end{axis}

\end{tikzpicture}
    \caption{Cumulative regret of our algorithm in log scale. For reference purpose, we also plot $\sqrt{T}$ and $T$ in log scale.   
    }
    \label{fig:cumu_reg_loglog}
\end{figure}
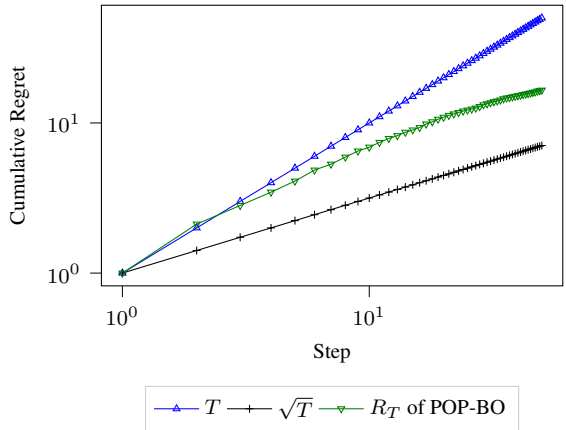

Fig.~\ref{fig:cumu_reg_loglog} shows the cumulative regret of POP-BO algorithm. The experimental conditions are the same as in Sec.~\ref{sec:exp_sampled_instance}. Note that both horizontal and vertical axes in Fig.~\ref{fig:cumu_reg_loglog} are in log scale, and thus the slope of the curve roughly represents the power of the cumulative regret. It can be clearly seen that the order of the cumulative regret is between $\sqrt{T}$ and $T$~(indeed, close to $T^{\frac{3}{4}}$ by checking the slope in log scale), which verifies our theoretical results in Thm.~\ref{thm:kern_spec_bound}. 

\section{Kernel-Specific Convergence Rate}
\label{app_sec:kern_spec_conv_rate}
Similar to the bounds in the Appendix~\ref{app_sec:kern_spec_Rbound}, we can plug in the kernel-specific covering number and maximum information gain to derive the kernel-specific convergence rate in Tab.~\ref{tab:kern_spec_bounds}.   
\begin{table*}[htbp]
    \centering
    \caption{Kernel-specific convergence rate for $x_{t^\star}$.}
    \label{tab:kern_spec_bounds}
    {\begin{tabular}{|c|c|c|c|}
      \hline 
       Kernel &  Linear & Squared Exponential & Mat\'ern $\left(\nu>\frac{d}{4}(3+d+\sqrt{d^2+14d+17})=\Theta(d^2)\right)$\\
       \hline 
       $f(x^\star)-f(x_{t^\star})$ & ${\mathcal{O}}\left(\frac{(\log T)^{\nicefrac{3}{4}}}{T^{\nicefrac{1}{4}}}\right)$  & ${\mathcal{O}}\left(\frac{(\log T)^{\nicefrac{3}{4}(d+1)}}{T^{\nicefrac{1}{4}}}\right)$ &${\mathcal{O}}\left(\frac{(\log T)^{\nicefrac{3}{4}}T^{\frac{d}{\nu}\left(\frac{1}{4}+\frac{d+1}{4+2(d+1)\nicefrac{d}{\nu}}\right)}}{T^{\nicefrac{1}{4}}}\right)$ \\
       \hline 
    \end{tabular}
    }
\end{table*}

\section{More Experimental Results and Details}
\label{app_sec:exp_res_detail}
\textbf{Selection of Hyperparameters}. Three key hyperparameters that influence the performance of POP-BO are the kernel lengthscale, the norm bound and the confidence level term $\beta$ as shown in Thm.~\ref{thm:conf_set_f}. We set $\beta=\beta_0\sqrt{t}$, where $\beta_0$ is set to $1.0$ by default. For the sampled instances from Gaussian processes, the lengthscale is set to be the ground truth and the norm bound is set to be $1.1$ times the ground truth. For the test function examples, we choose the lengthscale by maximizing the likelihood value over a set of randomly sampled data and set the norm bound to be $6$ by default (with the test functions all normalized).

\textbf{Details on Sampled Instances from Gaussian Process}. Specifically, we randomly sample some knot points from a joint Gaussian distribution marginalized from the Gaussian process, and then construct its corresponding minimum-norm interpolant~\cite{maddalena2021deterministic} as the ground truth function. 

\textbf{Empirical Method for Reporting a Solution}.
In the experiment of test function optimization, we report the point that maximizes the minimum norm maximum likelihood estimator $\hat{f}^{\textrm{MLE}}_t$, which achieves better empirical performance.

\rev{
\textbf{Solution Report Method for Baselines}.
The approach to reporting a solution is the same as in the original paper of the baseline algorithm if it is mentioned. Therefore, for the baseline qEUBO~\cite{astudillo2023qeubo}, we report the solution that maximizes the expected objective value conditioned on the historical samples. For the baseline SGP~\cite{pmlr-v202-takeno23b}, we report the first point of the duel proposed by the algorithm in step $t$. For the baseline DTS~\cite{gonzalez2017preferential}, we report the Condorcet winner. 
}

\rev{
\textbf{Effect of Hyperparameters}. We conducted more experiments to assess the effect of hyperparameters. We observe that the hyperparameters with most influence are the norm bound $B$ and the confidence level $\beta_t$. The larger the norm bound $B$ is, the more variance the estimate function has. If $B$ is set too large, the convergence for the suboptimality of the reported solution tends to be slower. $\beta_t$ can be set to be $\beta_0\sqrt{t}$ in practice and determines the level of exploration, where $\beta_0$ is a fixed constant. The larger $\beta_0$ is, the more explorative the algorithm is and may have higher cumulative regret. But setting $\beta_0$ to be very small may also cause weak exploration and make the suboptimality of the reported solution converge slower. 

}

\rev{
\subsection{Experimental Results for Higher-Dimensional Problems}
\label{app_sec:high_dim_res}
\subsubsection{Higher-Dimensional Problems Sampled from Gaussian Process}
We consider the optimization of $7$-dimensional black-box function sampled from a Gaussian process with kernel function as shown in Eq.~\meqref{eqn:high_dim_se_kern}, 
\begin{equation} 
\label{eqn:high_dim_se_kern}
k(x,\bar{x}) =\sigma_\mathrm{SE}^2\exp{\left\{-\frac{\norm{x-\bar{x}}^2}{l^2}\right\}}
\end{equation}
where $\sigma_\mathrm{SE}^2=9.0$ and $l=5\sqrt{7}$. The optimization domain is set to be $[0,10]^7$. We run $20$ randomly sampled instances for $100$ steps. The average update time for each step $t$ is only $11.0$ seconds on a personal computer with one \texttt{Intel64 Family 6 Model 142 Stepping 12 GenuineIntel ~1803 Mhz} processor and \texttt{16.0 GB RAM}. This is comparably very small considering that each query to the comparison oracle can be very expensive in practice (e.g., heating the room up to a certain temperature to evaluate occupant comfort, which may take tens of minutes). We compare our method to the SGP baseline. 
}


\begin{figure}[h!]
    \centering
    \includegraphics[width=0.73\columnwidth]{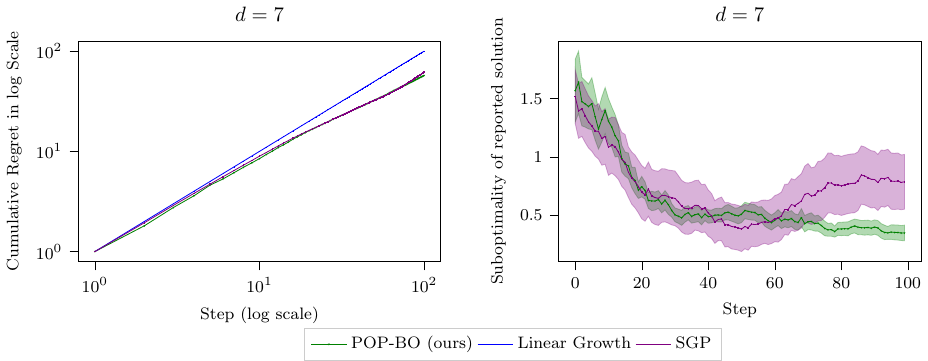}
    \caption{\rev{Cumulative regret in log scale and the suboptimality of reported solution in linear scale for a $7$-dimensional problem sampled from Gaussian process. For reference purpose, we also plot $T$ in the cumulative regret plot in log scale, where the shaded areas represent $\pm0.2\textsf{ standard deviation}$.}   
    }
    \label{fig:cumu_sim_reg_high_gp}
\end{figure}

\rev{
Fig.~\ref{fig:cumu_sim_reg_high_gp} shows the cumulative regret (in log scale) and the suboptimality of the reported solution for our POP-BO algorithm, where the reported solution is derived by maximizing the maximum likelihood estimate function. It can be clearly seen that our algorithm achieves both sublinear regret growth and fast convergence for the suboptimality of the reported solution in this 7-dimensional problem. Interestingly, the suboptimality of SGP converges similarly to our method before 50 steps, but get even worse after 50 steps. This is because SGP ignores the randomness in the preference feedback, which leads to misbelief in the function difference value, and such misbelief is more significant when the function difference value is small.  
}

\rev{

}

\rev{
\subsubsection{Higher-Dimensional Test Problem}
In this section, we further consider the optimization of the $6$-dimensional Ackley function as shown in~\cite{astudillo2023qeubo}. For this problem, we compare POP-BO algorithm to the qEUBO algorithm proposed in~\cite{astudillo2023qeubo}. Fig.~\ref{fig:cumu_sim_reg_ackley} shows the cumulative regret and the suboptimality of the reported solution. In this particular problem, qEUBO performs better than our POP-BO algorithm in terms of cumulative regret, while our POP-BO algorithm performs slightly better than qEUBO in terms of the suboptimality of the reported solution. 

\begin{figure}[h!]
    \centering   \includegraphics[width=0.73\columnwidth]{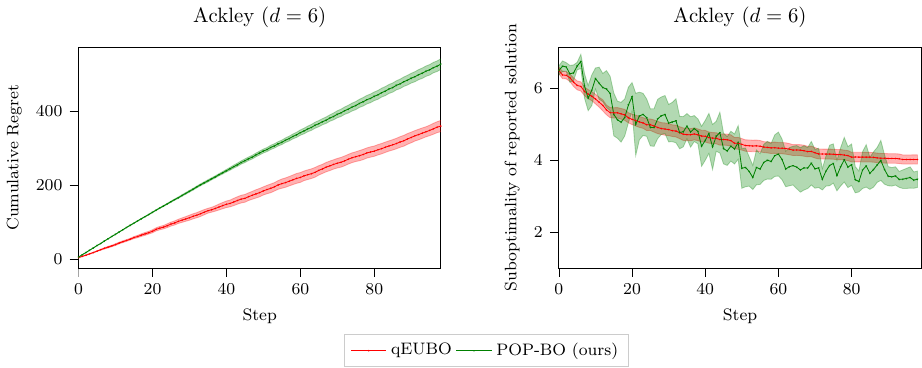}
    \caption{\rev{Cumulative regret and the suboptimality of reported solution for the $6$-dimensional Ackley function optimization problem, where the shaded areas represent $\pm0.5\textsf{ standard deviation}$.}   
    }
    \label{fig:cumu_sim_reg_ackley}
\end{figure}

}

\subsection{Occupant Thermal Comfort Optimization}
\subsubsection{Two-Dimensional Comfort Optimization}
\begin{figure}[thbp]
    \centering
    \includegraphics[width=0.73\columnwidth]{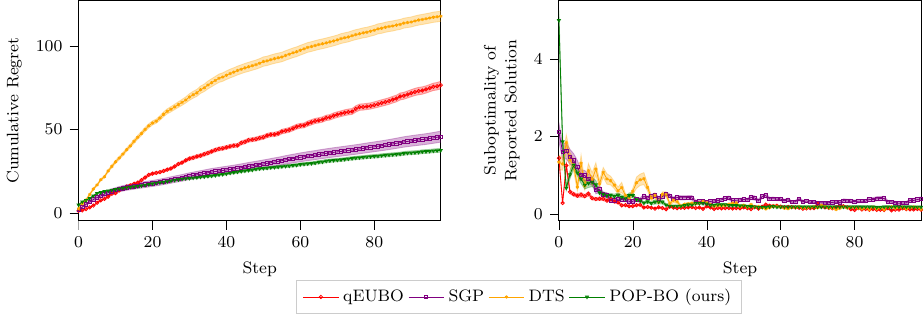}
    \caption{Cumulative regret and the suboptimality of reported solution of different algorithms for thermal comfort optimization.}
    \label{fig:thermal_comfort}
\end{figure}

An accurate model of human thermal comfort is crucial for improving occupants' comfort while saving energy in buildings. However, establishing such a model has proven to be a complex and challenging task \cite{zhang_bayesian_2024} and standard offline models ignore the individual differences among occupants. In this section, we consider the real-world problem of maximizing occupant thermal comfort directly from thermal preference feedback. To emulate real human thermal sensation, we use the well-known and widely adopted Predicted Mean Vote (PMV) model~\cite{fanger1970thermal} as the ground truth and generate the preference feedback according to the Bernoulli model as assumed in Assumption~\ref{assump:pref_model}. We optimize the indoor air temperature and air speed, which are the two major factors that influence thermal comfort and are controllable by HVAC (Heating, Ventilation, and Air Conditioning) systems and fans. Indeed, tuning these two factors has been proven effective in providing thermal comfort while minimizing energy consumption \cite{LYU2023111002}. The result is shown in Fig. \ref{fig:thermal_comfort} where the mean is taken over 30 instances of simulation. It can be seen that our method stably achieves superior performance in optimizing human thermal comfort, which implies its potential to deal with preferential feedback in real-world applications. It is also noticeable that although qEUBO achieves slightly better performance in terms of the convergence of the reported solution, the cumulative regret of qEUBO is almost twice of POP-BO's cumulative regret. This means our method is more favorable in applications where online performance during the optimization is also critical, such as online tuning of HVAC systems. 

\subsubsection{\rev{Scalability to higher dimension}}
\label{app_subsec:therm_scale_to_higher}
\begin{figure}[thbp]
    \centering
    \includegraphics[width=0.73\columnwidth]{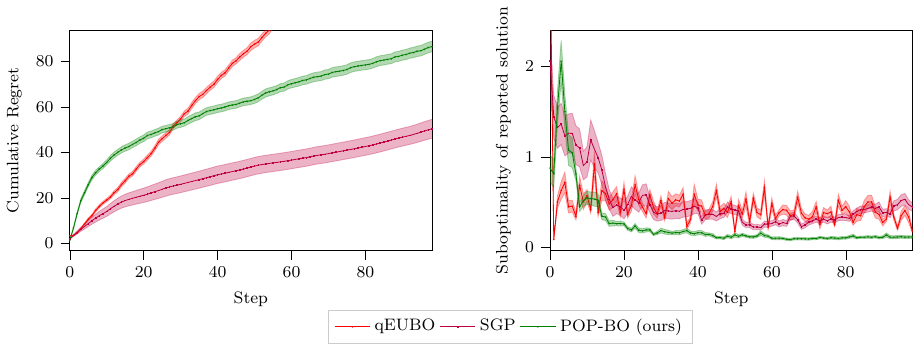}
    \caption{\rev{Cumulative regret and the suboptimality of reported solution of different algorithms for the four-dimensional thermal comfort optimization problem.}}
    \label{fig:thermal_comfort_4D}
\end{figure}

\rev{Additionally, to demonstrate the scalability of POP-BO in this real-world comfort optimization problem, we additionally tune the mean radiant temperature and relative humidity, which results in a four-dimensional black-box optimization problem. The result is shown in Fig. \ref{fig:thermal_comfort_4D}. It can be observed that increasing the dimensionality does not drastically decrease the convergence rate of our method. Furthermore, the baseline method qEUBO can decrease the objective value very fast in the initial steps, but seems to be still very oscillatory after 10 steps. In contrast, our method converges faster than SGP without the oscillation issue like qEUBO.}

\begin{figure}[thbp]
    \centering
\includegraphics[width=0.73\columnwidth]{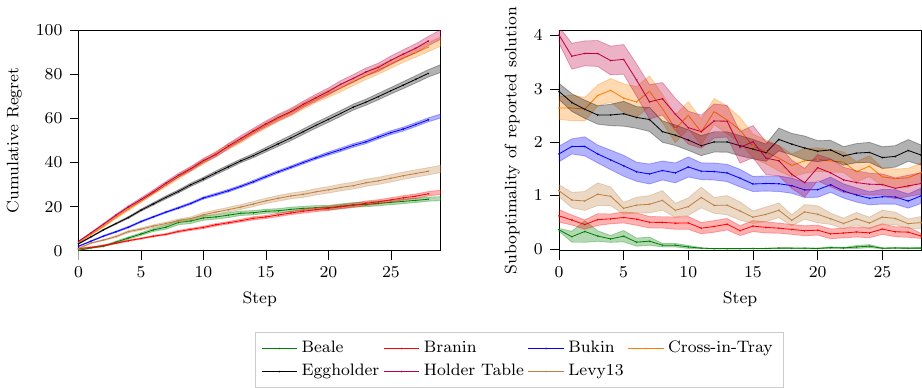}
    \caption{\rev{Cumulative regret and the suboptimality of reported solution of POP-BO algorithm for the different test problems in Tab.~\ref{tab:subopt_testfunc}.}}
    \label{fig:test_prob_popbo}
\end{figure}

\rev{
\subsection{Details About the Results in Tab.~\ref{tab:subopt_testfunc}}
The cumulative regret and evolution of suboptimality for the different test problems in Tab.~\ref{tab:subopt_testfunc} are shown in Fig.~\ref{fig:test_prob_popbo}. Since the considered problems only have $2$-dimensional input and in the applications of Bayesian optimization, it is typically desired to obtain a set of solution with objective value as close to the optimal value as possible. So we only consider $30$ steps here. Other baselines can make limited progress in terms of the suboptimality of the reported solution within only $30$ steps (partially also due to the `adversarial' property of the test functions, i.e., severe non-convexity and multiple local maxima) as shown in Tab.~\ref{tab:subopt_testfunc}. To the sharp contrast, our POP-BO algorithm makes significant progress in reducing the suboptimality of the reported solution by balancing exploration and exploitation, and estimating the best solution in a principled way.     
}

\begin{figure}[thbp]
    \centering
\includegraphics[width=0.73\columnwidth]{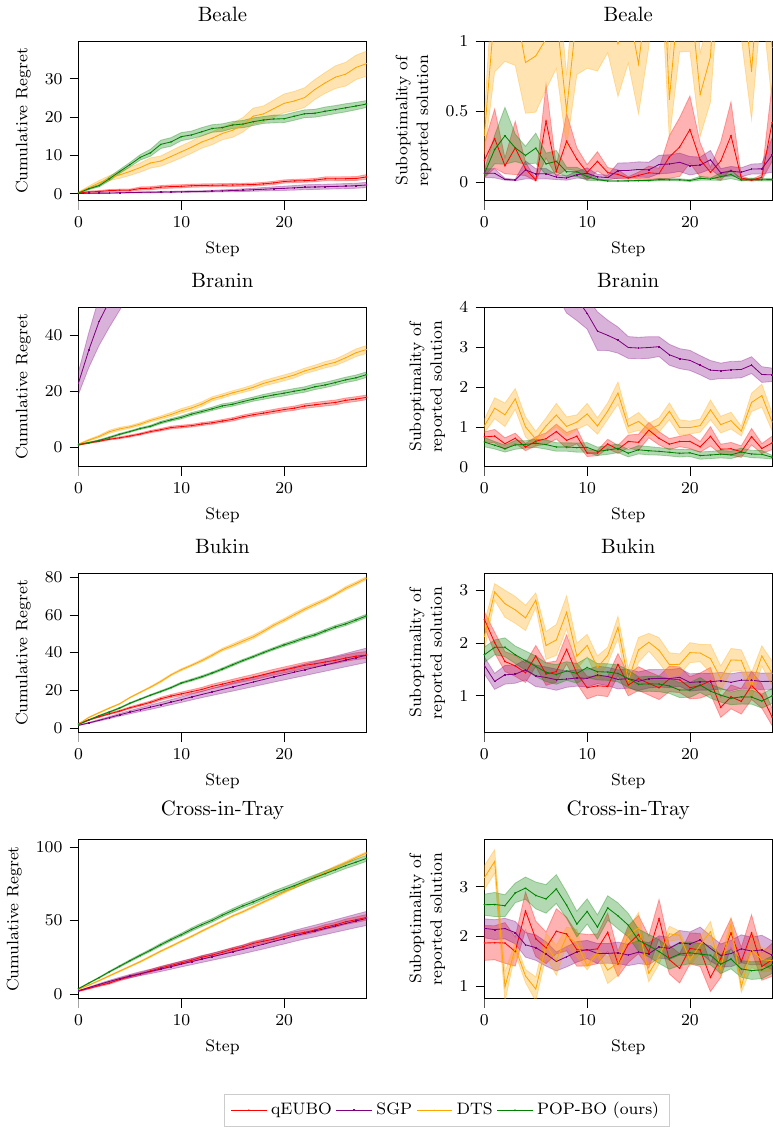}
    \caption{\rev{Cumulative regret and the suboptimality of reported solution of different algorithms for the test problems Beale, Branin, Bukin, and Cross-in-Tray in Tab.~\ref{tab:subopt_testfunc}.}}
    \label{fig:test_prob_group_1}
\end{figure}

\begin{figure}[thbp]
    \centering
\includegraphics[width=0.73\columnwidth]{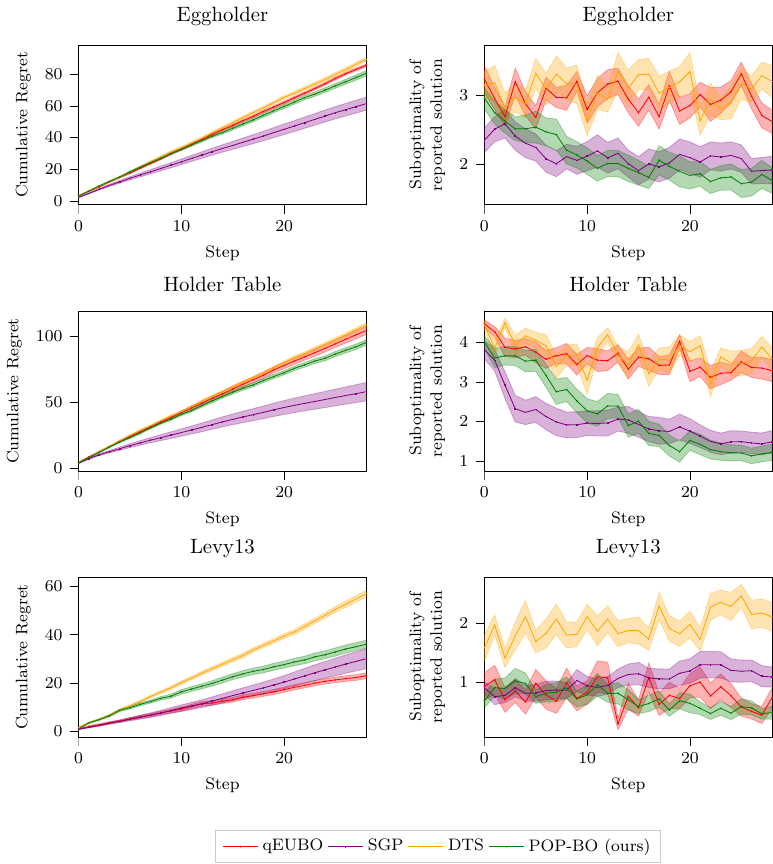}
    \caption{\rev{Cumulative regret and the suboptimality of reported solution of different algorithms for the test problems Eggholder, Holder Table, and Levy13 in Tab.~\ref{tab:subopt_testfunc}.}}
    \label{fig:test_prob_group_2}
\end{figure}

\rev{To provide more insights into POP-BO's performance across different settings, we compare our algorithm's evolution of cumulative regret and suboptimality to other baseline methods for each test problem in Fig.~\ref{fig:test_prob_group_1} and Fig.~\ref{fig:test_prob_group_2}. It can be observed that our method may perform slightly worse than some baselines in certain problems. For example, our method performs slightly worse than qEUBO in the Bukin problem in terms of suboptimality. However, our method performs stably and is consistently one of the best in all the test problems in terms of the suboptimality.       

}

\rev{
\section{Additional Contributions as Compared to \cite{mehta2023kernelized}}
\label{sec:add_contri}
Notably, \cite{mehta2023kernelized} proposes Borda-AE algorithm, which directly learns the winning probability function using kernel ridge regression. This key design allows the authors to derive an information-theoretic convergence rate and efficient computation method without diving into the learning of the underlying reward function. 

However, \cite{mehta2023kernelized} has key limitations and our paper makes additional contributions in the following two aspects.
\begin{enumerate}

    \item\textbf{Cumulative regret bound.} 
    There are two possible ways to define cumulative regret. One way is that we can define the (partial) cumulative regret as the summation of the suboptimality of \emph{only} $x_t$ (that is, $\sum_{t=1}^T(f(x^\star)-f(x_t))$). With this (partial) cumulative regret definition, Borda-AE algorithm can provide a sublinear (partial) cumulative regret bound, although it has linear growth in the cumulative regret of the compared point sequence $\{x_t^\prime\}_{t=1}^T$. However, in many practical online learning applications, it is desired to control the suboptimality of both $x_t$ and $x_t^\prime$ sequences. For example, when tuning the thermal/visual comfort of room occupants, we require the occupants to experience both $x_t$ and $x_t^\prime$ conditions for comparison purposes and the suboptimality (links to discomfort) caused by both $x_t$ and $x_t^\prime$ need to be managed.        

    Therefore, it is more practically relevant to define (total) cumulative regret as the total cumulative suboptimality of both $x_t$ and $x_t^\prime$ sequences (that is, $\sum_{t=1}^T(f(x^\star)-f(x_t))+\sum_{t=1}^T(f(x^\star)-f(x_t^\prime))$). Interestingly, since $x_t^\prime=x_{t-1}$ by the design of our POP-BO algorithm, this (total) cumulative regret bound reduces to $2\sum_{t=1}^T(f(x^\star)-f(x_t))$, for which we provide our sublinear cumulative regret bound. As such, the (total) cumulative regret bound provided by our paper is stronger than the (partial) cumulative regret bound that could be obtained by~\cite{mehta2023kernelized}.


    \item\textbf{Applicability to online learning problem.} 
    Following the last point, \cite{mehta2023kernelized} is not applicable to the online learning problem since in line 6 of the Borda-AE algorithm, $a_t^\prime$ is uniformly sampled from the action space, which leads to a linear growth of cumulative regret. This means Borda-AE has very poor online performance and can not be applied to an online learning problem. For example, in building thermal comfort tuning, we also want to control the discomfort caused during the tuning process. In contrast, our POP-BO algorithm has good online performance with both a theoretical bound on cumulative regret~(Thm. 5.2) and empirical evidence on small cumulative regret~(Fig.~2).      
\end{enumerate}
}
\end{document}